\documentclass[sigconf]{acmart}

\AtBeginDocument{%
  }

\copyrightyear{2024}
\acmYear{2024}
\setcopyright{rightsretained}
\acmConference[MM '24]{Proceedings of the 32nd ACM International Conference on Multimedia}{October 28-November 1, 2024}{Melbourne, VIC, Australia}
\acmBooktitle{Proceedings of the 32nd ACM International Conference on Multimedia (MM '24), October 28-November 1, 2024, Melbourne, VIC, Australia}
\acmDOI{10.1145/3664647.3681046}
\acmISBN{979-8-4007-0686-8/24/10}

\usepackage[utf8]{inputenc}
\usepackage{stfloats}
\usepackage{pifont} 
\usepackage{algorithm}
\usepackage[noend]{algpseudocode}
\usepackage{url}
\usepackage{gensymb}
\usepackage{hyperref}
\hypersetup{hidelinks,
	colorlinks=true,
	allcolors=black,
	pdfstartview=Fit,
	breaklinks=true}

\begin{document}

\title{Freehand Sketch Generation from Mechanical Components}


% \affiliation{%

\author{Zhichao Liao}
\authornote{Co-first authors. Listing order is random.}
\email{liaozc23@mails.tsinghua.edu.cn}
\orcid{0009-0004-9125-9428}
\affiliation{%
  \institution{Tsinghua Shenzhen International Graduate School, Tsinghua University}
  \city{Shenzhen}
  \country{China}
}

\author{Di Huang}
\authornotemark[1]
\email{huangdi@sz.tsinghua.edu.cn}
\orcid{0009-0009-7336-161X}
\affiliation{%
  \institution{Tsinghua Shenzhen International Graduate School, Tsinghua University}
  \city{Shenzhen}
  \country{China}
  }

\author{Heming Fang}
\authornotemark[1]
\email{22225131@zju.edu.cn}
\orcid{0009-0001-0902-0441}
\affiliation{%
  \institution{State Key Laboratory of Fluid Power \& Mechatronic Systems, Zhejiang University}
  \city{Hangzhou}
  \country{China}
  }

\author{Yue Ma}
\email{y-ma21@mails.tsinghua.edu.cn}
\orcid{0009-0000-5485-4237}
\affiliation{%
  \institution{Tsinghua Shenzhen International Graduate School, Tsinghua University}
  \city{Shenzhen}
  \country{China}
  }

\author{Fengyuan Piao}
\authornote{Corresponding author.}
\email{pfy22@mails.tsinghua.edu.cn}
\orcid{0009-0005-6688-8916}
\affiliation{%
  \institution{Tsinghua Shenzhen International Graduate School, Tsinghua University}
  \city{Shenzhen}
  \country{China}
  }

\author{Xinghui Li}
\authornotemark[2]
\email{li-xh21@mails.tsinghua.edu.cn}
\orcid{0009-0009-9813-6259}
\affiliation{%
  \institution{Tsinghua Shenzhen International Graduate School, Tsinghua University}
  \city{Shenzhen}
  \country{China}
  }

\author{Long Zeng}
\authornotemark[2]
\email{zenglong@sz.tsinghua.edu.cn}
\orcid{0000-0002-3090-6319}
\affiliation{%
  \institution{Tsinghua Shenzhen International Graduate School, Tsinghua University}
  \city{Shenzhen}
  \country{China}
  }

\author{Pingfa Feng}
\email{fengpf@tsinghua.edu.cn}
\orcid{0000-0002-8090-1508}
\affiliation{%
  \institution{Tsinghua Shenzhen International Graduate School, Tsinghua University}
  \city{Shenzhen}
  \country{China}
  }

% \affiliation{%

% \affiliation{%

% \affiliation{%

% \affiliation{%

% \affiliation{%

% \affiliation{%

% \affiliation{%

\renewcommand{\shortauthors}{Zhichao Liao et al.}

\begin{abstract}
Drawing freehand sketches of mechanical components on multimedia devices for AI-based engineering modeling has become a new trend. However, its development is being impeded because existing works cannot produce suitable sketches for data-driven research. These works either generate sketches lacking a freehand style or utilize generative models not originally designed for this task resulting in poor effectiveness. To address this issue, we design a two-stage generative framework mimicking the human sketching behavior pattern, called MSFormer, which is the first time to produce humanoid freehand sketches tailored for mechanical components. The first stage employs Open CASCADE technology to obtain multi-view contour sketches from mechanical components, filtering perturbing signals for the ensuing generation process. Meanwhile, we design a view selector to simulate viewpoint selection tasks during human sketching for picking out information-rich sketches. The second stage translates contour sketches into freehand sketches by a transformer-based generator. To retain essential modeling features as much as possible and rationalize stroke distribution, we introduce a novel edge-constraint stroke initialization. Furthermore, we utilize a CLIP vision encoder and a new loss function incorporating the \textit{Hausdorff distance} to enhance the generalizability and robustness of the model. Extensive experiments demonstrate that our approach achieves state-of-the-art performance for generating freehand sketches in the mechanical domain. Project page: \href{https://mcfreeskegen.github.io/}{\textit{\textcolor{cyan}{https://mcfreeskegen.github.io/}}}.
\end{abstract}


% \keywords{Do, Not, Us, This, Code, Put, the, Correct, Terms, for,




\begin{teaserfigure}
  \centering
  \includegraphics[width=0.82\textwidth,
  height=0.351\textwidth
  ]{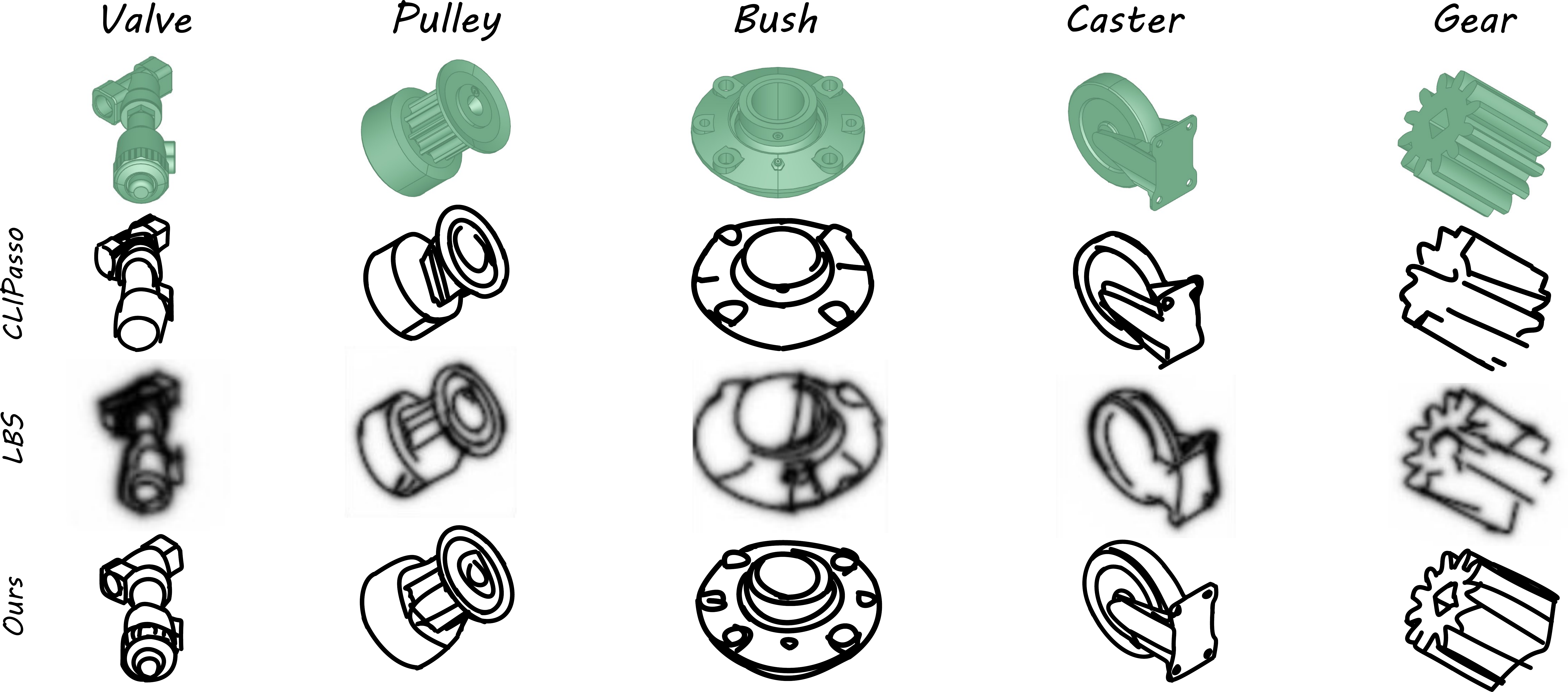}
  \caption{Various mechanical freehand sketches generated by ours and other approaches. Our method produces sketches from mechanical components while maintaining a freehand style and their essential modeling features, e.g., pulley grooves, through holes on the bush, and gear teeth. }
  \Description{}
  \label{fig:teaser}
\end{teaserfigure}

\maketitle

\section{Introduction}

Nowadays, with the vigorous development of multimedia technology, a new mechanical modeling approach has gradually emerged, known as freehand sketch modeling \cite{10.1145/3528223.3530133,10.1145/3414685.3417807} . Different from traditional mechanical modeling paradigms \cite{ZENG20121115} , freehand sketch modeling on multimedia devices does not require users to undergo prior training with CAD tools. In the process of freehand sketch modeling for mechanical components, engineers can utilize sketches to achieve tasks such as component sketch recognition, components fine-grained retrieval based on sketches \cite{ZENG201982} , and three-dimensional reconstruction from sketches to components. 
Modeling in this way greatly improves modeling efficiency. 
However, limited by the lack of appropriate freehand sketches for these data-driven studies in the sketch community, the development of freehand sketch modeling for mechanical components is hindered. It is worth emphasizing that manual sketching and collecting mechanical sketches is a time-consuming and resource-demanding endeavor. To address the bottleneck, we propose a novel two-stage generative model to produce freehand sketches from mechanical components automatically.

To meet the requirements of information richness and accuracy for modeling, we expect that freehand sketches used for mechanical modeling maintain a style of hand-drawn while preserving essential model information as much as possible. Previous works that generate engineering sketches
\cite{manda2021cadsketchnet,han2020spare3d,NEURIPS2021_28891cb4,seff2020sketchgraphs,willis2021engineering} , primarily focus on perspective and geometric features of models. As a result, their sketches lack a hand-drawn style, making them unsuitable as the solution of data generation for freehand sketch modeling. Existing data-driven freehand sketch generation methods \cite{ge2020creative,liu2020neural,zhang2015end,tong2021sketch,bhunia2022doodleformer,cao2019ai,luhman2020diffusion,wang2022sketchknitter,Ribeiro_2020_CVPR,lin2020sketch} also fall short in this task because they require the existence and availability of relevant datasets. While CLIPasso \cite{vinker2022clipasso} and LBS \cite{lee2023learning} can produce abstract sketches without additional datasets, as shown in Figure \ref{fig:teaser} , their results for mechanical components are afflicted by issues such as losing features, line distortions, and random strokes. In contrast, we propose a mechanical vector sketch generation technique that excels in maintaining  precise and abundant modeling features and a freehand style without additional sketch datasets.

Our method, the first time to generate freehand sketches for mechanical components, employs a novel two-stage architecture. It mimics the human sketching behavior pattern which commences with selecting optimal viewpoints, followed by hand-sketching. In Stage-One, we generate multi-perspective contour sketches from mechanical components via Open CASCADE, removing irrelevant information for engineering modeling which may also mislead stroke distribution in generated sketches. To select information-rich sketches, we devise a view selector to simulate the viewpoint choices made by engineers during sketching. Stage-Two translates regular contour sketches into humanoid freehand sketches by a transformer-based generator. It is trained by sketches created by 
a guidance sketch generator that utilizes our innovative edge-constraint initialization to retain more modeling features. 
Our inference process relies on trained weights to stably produce sketches defined as a set of Bézier curves. 
Additionally, we employ a CLIP vision encoder combining a pretrained vision transformer \cite{dosovitskiy2020image} ViT-B/32 model of CLIP \cite{CLIP} with an adapter \cite{gao2024clip} , which utilizes a self-attention mechanism \cite{vaswani2017attention} to establish global relations among graph blocks, enhancing the capture of overall features. It fortifies the method's generalization capability for unseen models during training and inputs with geometric transformation (equivariance). Furthermore, our proposed new guidance loss, incorporating the \textit{Hausdorff distance}, considers not only the spatial positions but also the boundary features and structural relationships between shapes. It improves model's ability to capture global information leading to better equivariance. Finally, we evaluate our method both quantitatively and qualitatively on the collected mechanical component dataset, which demonstrates the superiority of our proposed framework. We also conduct ablation experiments on key modules to validate their effectiveness.

In summary, our contributions are the following:

% \vspace{-2em}

\begin{itemize}
\item As far as our knowledge goes, this is the first time to produce freehand sketches tailored for mechanical components. To address this task, we imitate the human  sketching  behavior pattern to design a novel  two-stage sketch generation framework. 

 \item We introduce an innovative edge-constraint initialization method to optimize strokes of guidance sketches, ensuring that outcomes retain essential modeling features and rationalize stroke distribution.

\item We utilize an encoder constituted by CLIP ViT-B/32 model and an adapter to improve the generalization and equivariance of the model. Furthermore, we propose a novel \textit{Hausdorff distance}-based guidance loss to capture global features of sketches, enhancing the method's equivariance.

\item Extensive quantitative and qualitative experiments demonstrate that our approach can achieve state-of-the-art performance compared to previous methods. 

\end{itemize}

\section{Related Work}
Due to little research on freehand sketch generation from mechanical components, there is a review of mainstream generation methods relevant to our work in the sketch community.

\noindent{\bfseries Traditional Generation Method } In the early stages of sketch research, sketches from 3D models were predominantly produced via traditional edge extraction methodologies \cite{sobel19683x3,canny1986computational,winnemoller2012xdog,prewitt1970object,marr1977analysis,manda2021cadsketchnet} . Among them, Occluding contours \cite{marr1977analysis} which detects the occluding boundaries between foreground objects and the background to obtain contours, is the foundation of non-photorealistic 3D computer graphics. Progressions in occluding contours \cite{marr1977analysis} have catalyzed advancements in contour generation, starting with Suggestive contours \cite{suggestive} , and continuing with Ridge-valley lines \cite{ohtake2004ridge} and kinds of other approaches \cite{apparent,manda2021cadsketchnet} . A comprehensive overview \cite{decarlo2012depicting} is available in existing contour generalizations. Similarly to the results of generating contours, Han et al. \cite{han2020spare3d} present an approach to generate line drawings from 3D models based on modeling information. Building upon previous work that solely focused on outlines of models, CAD2Sketch \cite{hahnlein2022cad2sketch} addresses the challenge of representing line solidity and transparency in results, which also incorporates certain drawing styles. However, all of these traditional approaches lack a freehand style like ours.

\noindent{\bfseries Learning Based Methods } Coupled with deep learning \cite{huang2024magicfight,ma2024followyourpose,ma2022simvtp,ma2022visual,ma2023magicstick,ma2024followyourclick,ma2024followyouremoji} , sketch generation approaches \cite{ge2020creative,liu2020neural,zhang2015end,tong2021sketch,bhunia2022doodleformer,cao2019ai,NEURIPS2021_28891cb4,8851809} have been further developed. Combining the advantage of traditional edge extraction approaches for 3D models and deep learning, Neural Contours \cite{liu2020neural} employs a dual-branch structure to leverage edge maps as a substitution for sketches. SketchGen \cite{NEURIPS2021_28891cb4} , SketchGraphs \cite{seff2020sketchgraphs} , and CurveGen and TurtleGen \cite{willis2021engineering} produce engineering sketches for Computer-Aided Design. However, such approaches generate sketches that only emphasize the perspective and geometric features of models, which align more closely with regular outlines, the results do not contain a freehand style. Generative adversarial networks (GANs) \cite{goodfellow2014generative} provide new possibilities for adding a freehand style to sketches \cite{ge2020creative,liu2020unsupervised,manushree2021xci,li2020artpdgan,wang2020robocodraw} . These approaches are based on pixel-level sketch generation, which is fundamentally different from how humans draw sketches by pens, resulting in unsuitability for freehand sketch modeling. Therefore, sketches are preferred to be treated as continuous stroke sequences. Recently, Sketch-RNN \cite{sketchRNN} based on recurrent neural networks (RNNs) \cite{RNNs} and variational autoencoders (VAEs) \cite{VAE} , reinforcement learning \cite{xie2013artist,zheng2018strokenet,pmlr-v80-ganin18a} , diffusion models \cite{huang2024diffusion,luhman2020diffusion,wang2022sketchknitter}
are explored for generating sketches. However, they perform poorly in generating mechanical sketches with a freehand style due to the lack of relevant training datasets. 
Following the integration of Transformer \cite{vaswani2017attention}  architectures into the sketch generation, the sketch community has witnessed the emergence of innovative models \cite{willis2021engineering,Ribeiro_2020_CVPR,lin2020sketch} . CLIPasso \cite{vinker2022clipasso} provides a powerful image to abstract sketch model based on CLIP \cite{CLIP} to generate vector sketches, but this method will take a long time to generate a single sketch. More critically, CLIPasso \cite{vinker2022clipasso} initializes strokes by sampling randomly, and optimizes strokes by using an optimizer for thousands of steps rather than based on pre-trained weights, leading to numerical instability. Despite LBS \cite{lee2023learning} being an improvement over Clipasso \cite{vinker2022clipasso} , it performs unsatisfactorily in generalization capability for inputs unseen or transformed. Compared to many previous approaches, our proposed generative model can produce vector sketches based on mechanical components, persevering key modeling features and a freehand style, greatly meeting the development needs of freehand sketch modeling.

\section{Method}
We first elaborate on problem setting in section \ref{3.1} . Then, we introduce our sketch generation process that presents Stage-One (CSG) and Stage-Two (FSG) of  MSFormer in sections \ref{3.2} and \ref{3.3} .

\subsection{Problem Setting}\label{3.1}
\begin{figure*}
  \includegraphics[width=\textwidth]{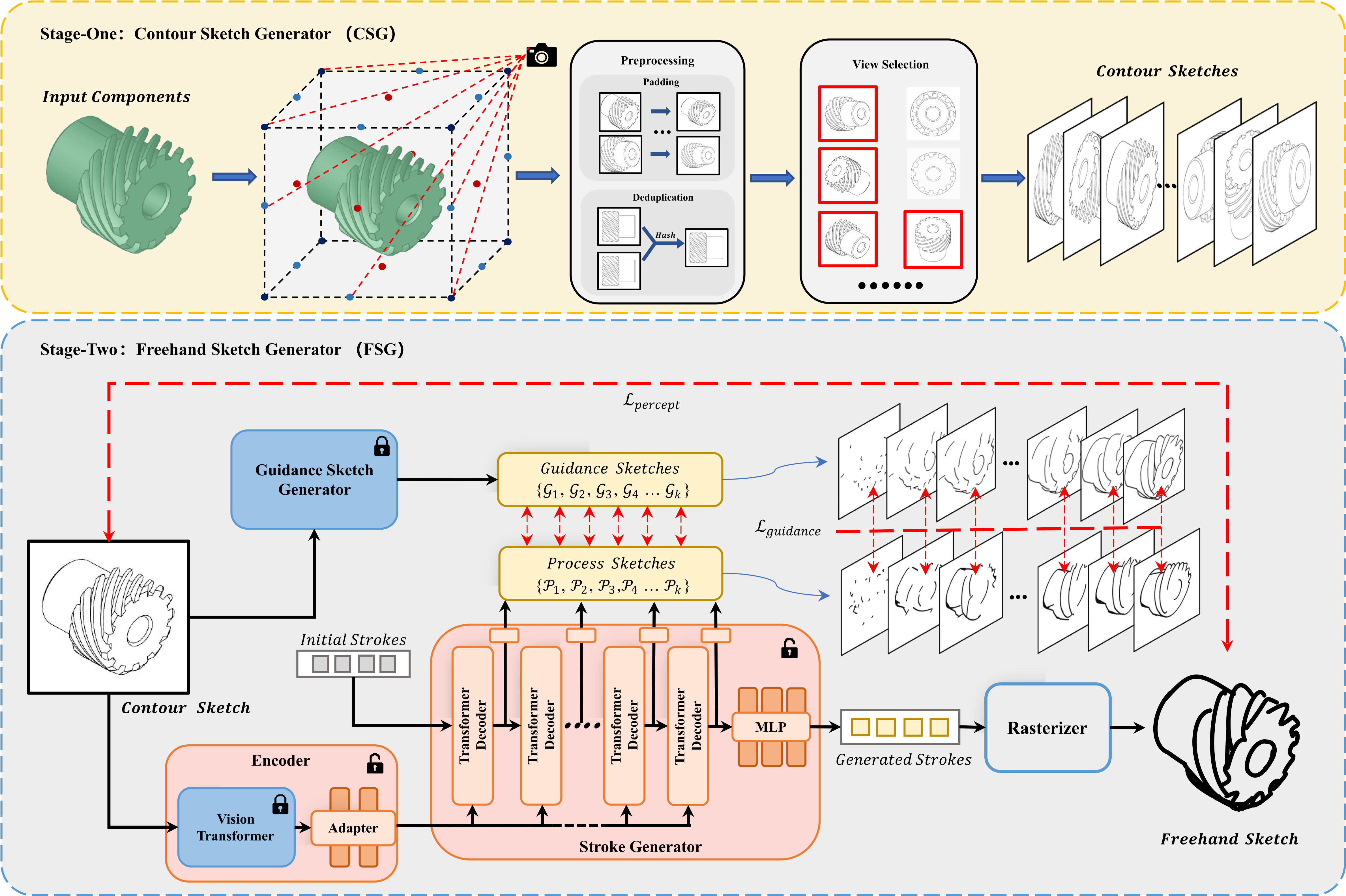}
  \caption{An overview of our method. (1) Stage-One: we generate contour sketches based on 26 viewpoints (represented by colorful points) of a cube (grey) . After that, Preprocessing and View Selection export appropriate contour sketches. 
  % from viewpoints which have significant features for freehand sketch modeling.
  (2) Stage-Two: By receiving initial strokes and features captured by our encoder from regular contour sketch, the stroke generator produces a set of strokes, which are next fed to a differentiable rasterizer to create a vector freehand sketch.
  % A vision transformer (ViT-B/32 model of CLIP) combined with two layers of adapters is used as the encoder to extracts the feature of contour sketch. A stroke generator based on transformer decoder optimizes initial strokes by progressive optimization process that process sketches $\{\mathcal{P}_1, \mathcal{P}_2, \ldots, \mathcal{P}_K\}$  from each intermediate layer are guided by guidance sketches$\{\mathcal{G}_1, \mathcal{G}_2, \ldots, \mathcal{G}_K\}$. Finally, We feed generated strokes to a differentiable rasterizer to produce a rasterized sketch.
  }
  \Description{}
  \label{fig:framework}
\end{figure*}

Given a mechanical component, our goal is to produce a freehand sketch. As depicted in Figure \ref{fig:framework} , it is carried out by stages: contour sketch generator and freehand sketch generator. We describe an   mechanical component as $\mathcal{M} \in {\Delta}^{3}$, where ${\Delta}^{3}$ represents 3D homogeneous physical space. Each point on  model corresponds to a coordinate $ ({x}_{i}, {y}_{i}, {z}_{i}) \in \mathbb{R}^{3} $, where $\mathbb{R}$ is information dimension. 
Through an affine transformation, a 3D model is transformed into 2D contour sketches $ \mathcal{C} \in {\Delta}^{2} $, which consists of a series of black curves expressed by pixel coordinates $ ({x}_{i}, {y}_{i})  \in \mathbb{R}^{2} $. In the gradual optimization process of Stage-Two, process sketches $\{\mathcal{P}_{i}\}_{i=1}^{K}$  are guided by guidance sketches $\{\mathcal{G}_{i}\}_{i=1}^{K}$, K is the number of sketches. Deriving from features of contour sketch $ \mathcal{C}$ and guidance sketches $\{\mathcal{G}_{i}\}_{i=1}^{K}$, our model produces an ultimate output freehand sketch $ \mathcal{S} $, which is defined as a set of n two-dimensional Bézier curves $\{s_1, s_2, \ldots, s_n\}$. Each of curve strokes is composed  by four control points  ${s}_{i}=\{(x_1,y_1)^{(i)},(x_2,y_2)^{(i)},(x_3,y_3)^{(i)},(x_4,y_4)^{(i)}\} \in \mathbb{R}^{8}, \forall i\in n $.

\subsection{Stage-One: Contour Sketch Generator}\label{3.2}
Contour Sketch Generator (CSG), called Stage-One, is designed for filtering noise (colors, shadows, textures, etc.) and simulating the viewpoint selection during human sketching to obtain recognizable and informative contour sketches from mechanical components. Previous methods optimize sketches using details such as the distribution of different colors and variations in texture. However, mechanical components typically exhibit monotonic colors and subtle texture changes. We experimentally observe that referencing this information within components not only fails to aid inference but also introduces biases in final output stroke sequences, resulting in the loss of critical features. As a result, when generating mechanical sketches, the main focus is on utilizing the contours of components to create modeling features.

Modeling engineers generally choose specific perspectives for sketching rather than random ones, such as three-view (Front/Top/ Right views), isometric view (pairwise angles between all three projected principal axes are equal), etc. As shown in Figure \ref{fig:framework} Stage-One, we can imagine placing a mechanical component within a cube and selecting centers of the six faces, midpoints of the twelve edges, and eight vertices of the cube as 26 viewpoints. Subsequently, we use PythonOCC \cite{paviot2018pythonocc} , a Python wrapper for the CAD-Kernel OpenCASCADE, to infer engineering modeling information  and render regular contour sketches of the model from these 26 viewpoints. 

Generated contour sketches are not directly suitable for subsequent processes. By padding, we ensure all sketches are presented in appropriate proportions. Given that most mechanical  components exhibit symmetry, the same sketch may be rendered from different perspectives. We utilize ImageHash technology for deduplication. Additionally, not all of generated sketches are useful and information-rich for freehand sketch modeling. For instance, some viewpoints of mechanical components may represent simple or misleading geometric shapes that are not recognizable nor effective for freehand sketch modeling. Therefore, we design a viewpoint selector based on ICNet \cite{zhao2018icnet} , which is trained by excellent viewpoint sketches picked out by modeling experts, to simulate the viewpoint selection task engineers face during sketching, as shown in Figure \ref{fig:framework} . Through  viewpoint selection, we obtained several of the most informative and representative optimal contour sketches for each mechanical component. 
The detailed procedure of Stage-One is outlined in Algorithm \ref{alg1}.

\begin{algorithm} 
\caption{Stage-One: Contour Sketch Generation}
\begin{flushleft}
\textbf{Input:} Mechanical components\\
\textbf{Output} Contour Sketches of mechanical components
\end{flushleft}
\begin{algorithmic}[1]
\Procedure{cad\_to\_contours}{}      
    \State $I \leftarrow$ Read a mechanical component in STEP format
    \State Set OCC to HLR mode and enable anti-aliasing
    \State $V1 \leftarrow$ Acquire contour sketches of $I$ from the 26 built-in viewpoints in OCC
    \State $V2 \leftarrow$ Center the object in $V1$ and maintain a margin from edges of the picture
    \State $V3 \leftarrow$ Remove duplicates from sketches in $V2$ using the $ImageHash$ library
    \State $O \leftarrow$ Filter the top $N$ contours with the most information from $V3$ using an image complexity estimator
\EndProcedure
\end{algorithmic}
\label{alg1}
\end{algorithm}

% \vspace{-1.5em}
\subsection{Stage-Two: Freehand Sketch Generator}\label{3.3}
Stage-Two, in Figure \ref{fig:framework} , comprises the Freehand Sketch Generator (FSG), which aims to generate freehand sketches based on regular contour sketches obtained from Stage-One. To achieve this goal, we design a transformers-based \cite{Ribeiro_2020_CVPR,Liu_2021_ICCV,lee2023learning} generator trained by guidance sketches, which stably generates freehand sketches with precise geometric modeling information. Our generative model does not require additional datasets for training. All training data are derived from the excellent procedural sketches produced by the guidance sketch generator. 

\noindent{\bfseries Generative Process } As illustrated in Figure \ref{fig:framework} , freehand sketch generator consists of four components: an encoder, a stroke generator, a guidance sketch generator, and a differentiable rasterizer. Our encoder utilizes CLIP ViT-B/32 \cite{CLIP} and an adapter to extract essential vision and semantic information from input. Although, in previous works, CLIPasso \cite{vinker2022clipasso} performs strongly in creating abstract sketches, it initializes strokes by sampling randomly and uses an optimizer for thousands of steps to optimize sketches, resulting in a high diversity of outputs and numerical instability. To a ensure stable generation of sketches, we design a training-based stroke generator that employs improved CLIPasso \cite{vinker2022clipasso} from the guidance sketch generator as ideal guidance. It allows us to infer high-quality sketches stably by utilizing pre-trained weights. Our stroke generator consists of eight transformer decoder layers and two MLP decoder layers. During training, to guarantee the stroke generator learns features better, process sketches $\{\mathcal{P}_{i}\}_{i=1}^{K}$ (K=8 in this paper) extracted from each intermediate layer are guided by guidance sketches $\{\mathcal{P}_{i}\}_{i=1}^{K}$  generated at the corresponding intermediate step of the optimization process in the guidance sketch generator. In the inference phase, the stroke generator optimizes initial strokes generated from trainable parameters into a set of n Bezier curves $ \{ s_1, s_2, \ldots, s_n \} $. These strokes are then fed into the differentiable rasterizer $ \mathcal{R} $ to produce a vector sketch $ \mathcal{S} = \mathcal{R}(s_1, \ldots, s_n ) = \mathcal{R}(\{ (x_j, y_j)^{(1)} \}_{j=1}^{4}, \ldots, \{ (x_j, y_j)^{(n)} \}_{j=1}^{4}) $.

\begin{figure}[ht] 
  \centering
  \includegraphics[width=\linewidth]{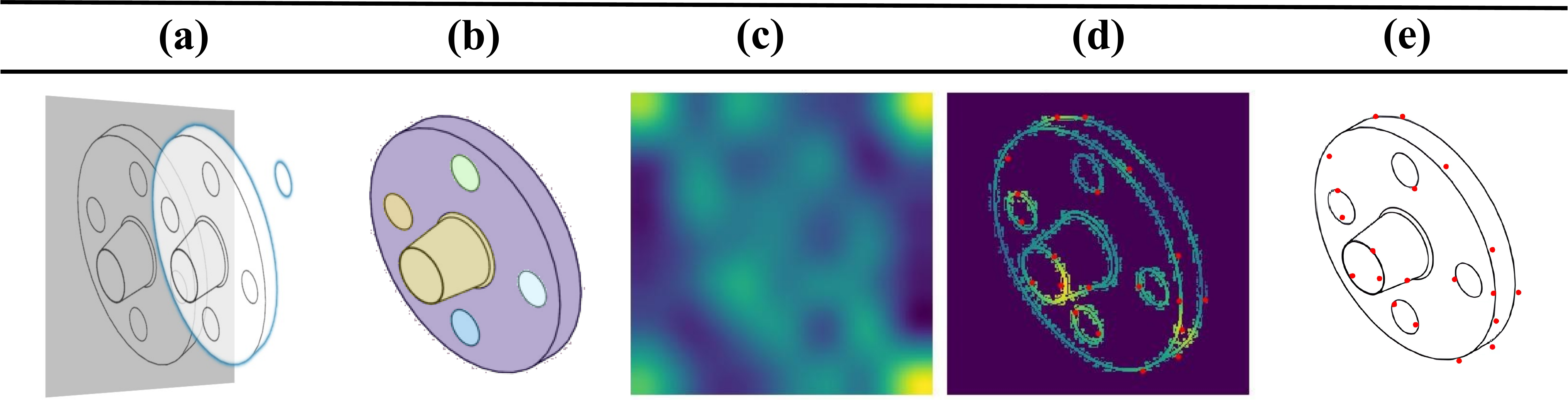}
  \caption{Edge-constraint Initialization. (a) and (b) are results of 
  segmenting through hole and overall segmentation  of flange by SAM \cite{kirillov2023segment} (distinguishing features through different coloring). (c) The saliency map generated from CLIP ViT activations. (d) and (e) are initial stroke locations (in red) in final distribution map and input. It is evident that our method accurately places initial strokes at features.}
  \label{fig:inital}
  
\end{figure}

\noindent{\bfseries Edge-constraint Initialization } 
The quality of guidance sketches plays a pivotal role in determining our outcomes' quality. Original CLIPasso \cite{vinker2022clipasso} initializes strokes via stochastic sampling from the saliency map. It could lead to the failure to accurately capture  features, as well as the aggregation of initial strokes in localized areas, resulting in generated stroke clutter. To address these issues, as shown in Figure \ref{fig:inital} , we modify the mechanism for initializing strokes 
in our guidance sketch generator. We segment contour sketches using SAM \cite{kirillov2023segment} and based on segmentation results accurately place the initial stroke on the edges of component's  features to constraint stroke locations. It ensures guidance generator not only generates precise geometric modeling information but also optimizes the distribution of strokes. Initialization comparison to original CLIPasso \cite{vinker2022clipasso} is provided in the   .

\noindent{\bfseries Encoder } FSG requires an encoder to capture features. Previous works for similar tasks predominantly employ a CNN encoder that solely relies on local receptive fields to capture features, making it susceptible to local variations and resulting in poor robustness for inputs unseen or transformed. While vision transformer (ViT) uses a self-attention mechanism \cite{vaswani2017attention} 
to establish global relationships between features. It enables the model to attend to overall information in inputs, unconstrained by fixed posture or shape. Therefore, we utilize  ViT-B/32 model of CLIP \cite{CLIP} to encode semantic understanding of visual depictions, which is trained on 400 million image-text pairs. And we combine it with 
an adapter that consists of two fully connected layers to fine-tune based on training data. As shown in Figure \ref{fig:comparsion-encoder} and Table \ref{tab:metrics evaluation} , our encoder substantially improves the robustness to unseen models during training and the equivariance.

\noindent{\bfseries Loss Function }
%=========================================================================PFY
 During training, we employ CLIP-based perceptual loss  to quantify the resemblance between generated freehand sketch $\mathcal{S}$ and contour sketch  $\mathcal{C}$ considering both geometric and semantic differences \cite{vinker2022clipasso,CLIP} . For synthesis of a sketch that is semantically similar to the given contour sketch, the goal is to constrict the distance in the embedding space of the CLIP model represented by $CLIP(x)$, defined as:
 % Leveraging the concept of semantic loss grounded on the CLIP model \cite{vinker2022clipasso}, our objective is to bridge the gap between the generated sketch $\textit{S}$ and the reference contour sketch $\textit{C}$ by evaluating both geometric and semantic disparities. For the synthesis of a sketch that resonates semantically with the given contour sketch, the goal is to constrict the distance in the embedding sphere of the CLIP model, articulated as:
\begin{equation}
   \mathcal{L}_{semantic} = \phi(CLIP(C), CLIP(S)),
\end{equation}
where $\phi$ represents the cosine proximity of the CLIP embeddings, \textit{i.e.},  $\phi(x,y) = 1 - \cos(x, y)$. Beyond this, the geometric similarity is measured by contrasting low-level features of output sketch and input contour, as follows:
\begin{equation}
    \mathcal{L}_{geometric} = \sum_{i=3,4}dist(CLIP_i(C), CLIP_i(S)),
\end{equation}
where $dist$ represents the $\mathcal{L}_{2}$ norm, explicitly, $dist(x, y) = \| x - y \|_2^2$, and $CLIP_i$ is the $i$-th layer CLIP encoder activation. As recommended by \cite{vinker2022clipasso}, we use layers 3 and 4 of the ResNet101 CLIP model. Finally, the perceptual loss is given by:
\begin{equation}
    \mathcal{L}_{percept} = \mathcal{L}_{geometric} + \beta_{s}\mathcal{L}_{semantic},
\end{equation}
\noindent where $\beta_{s}$ is set to 0.1. 
% =============================================================================

In the process of optimizing the stroke generator, a guidance loss is employed to quantify the resemblance between guidance sketches $\mathcal{G}$ and process sketches $\mathcal{P}$. Firstly, we introduce the \textit{Jonker-Volgenant algorithm} \cite{kuhn1955hungarian} to ensure that guidance loss is invariant to arrangement of each stroke's order, which is extensively utilized in assignment problems. The mathematical expression is as follows:
\begin{equation}
\mathcal{L}_{JK}=\sum_{k=1}^{K}\mathop{\min}_{\alpha}\sum_{i=1}^{n}\mathcal{L}_{1}(g_{k}^{(i)}, p_{k}^{\alpha(i)}), 
\end{equation}
where $ \mathcal{L}_1 $is the manhattan distance, $n$ is the number of strokes in the sketch. $ p_{k}^{(i)} $ is the $i$-th stroke of the sketch from the $k$-th middle process layer (with a total of $K$ layers), and $ g_{k}^{(i)} $ is the guidance stroke corresponding to $ p_{ k}^{\alpha(i)} $, $ \alpha $ is an arrangement of stroke indices.

Additionally, we innovatively integrate bidirectional \textit{Hausdorff distance} into the guidance loss, which serves as a metric quantifying the similarity between two non-empty point sets that our strokes can be considered as. It aids the model in achieving more precise matching of guidance sketch edges and maintaining structural relationships between shapes during training, thereby capturing more global features and enhancing the model's robustness to input with transformations. Experiment evaluation can be seen in section \ref{4.5} , The specific mathematical expression is as follows:
\begin{equation}
    \mathcal{\delta}_{H} = \max\{\tilde{\delta}_{H}(\mathcal{G},\mathcal{P}), \tilde{\delta}_{H}(\mathcal{P},\mathcal{G})\}, 
\end{equation}
where $\mathcal{P}=\{ p_{1},\ldots,p_{n} \}$ is the process sketch from each layer and $\mathcal{G}=\{ g_{1},\ldots,g_{n} \}$ is the guidance sketch corresponding to $\mathcal{P}$. $ g_{i} $ and $ p_{i} $ represent the strokes that constitute corresponding sketch. Both $ \mathcal{P} $ and $ \mathcal{G} $ are sets containing \textit{n} 8-dimensional vectors. $\tilde{\delta}_{H}(\mathcal{G},\mathcal{P})$ signifies the one-sided \textit{Hausdorff distance} from set $\mathcal{G}$ to set $\mathcal{P}$:
\begin{equation}
    \tilde{\delta}_{H}(\mathcal{G},\mathcal{P}) = \underset{g \in \mathcal{G}}{\max} \{\underset{p \in \mathcal{P}}{\min} \|g-p\|\},
\end{equation}
where $ \|\cdot\| $ is the Euclidean distance. Similarly, $\tilde{\delta}_{H}(\mathcal{P},\mathcal{G})$ represents the unidirectional \textit{Hausdorff distance} from set $\mathcal{P}$ to set $\mathcal{G}$:
\begin{equation}
    \tilde{\delta}_{H}(\mathcal{P},\mathcal{G}) = \underset{p \in \mathcal{P}}{\max} \{\underset{g \in \mathcal{G}}{\min} \|p-g\|\}.
\end{equation}

% The final guidance loss is as follows:
% \begin{equation}
%     \mathcal{L}_{guidance}=\lambda_{1}\mathcal{L}_{JK}+\lambda_{2}\mathcal{L}_{H},
% \end{equation}
% \noindent where $\lambda_{1}$ and $\lambda_{2}$ are the hyperparameters.

The guidance loss is as follows:
\begin{equation}
    \mathcal{L}_{guidance}=\mathcal{L}_{JK}+\beta_{h}\delta_{H},
\end{equation}
\noindent where $\beta_{h}$ is set to 0.8.

Our final loss function is as follows:
% \begin{equation}
%     \mathcal{L}_{toatl}={\lambda}_{p}\mathcal{L}_{percept}+{\lambda}_{g}\mathcal{L}_{guidance},
% \end{equation}
% \noindent with $ {\lambda}_{p}   $ and $ {\lambda}_{g}  $ are  hyperparameters.
\begin{equation}
    \mathcal{L}_{total}=\mathcal{L}_{percept}+\mathcal{L}_{guidance}.
\end{equation}

\section{Experiments}
\subsection{Experimental Setup} \label{4.1}
% Detailed information about each experiment is provided in the Appendix.
\noindent{\bfseries Dataset } We collect mechanical components in STEP format from  TraceParts \cite{TraceParts} databases, encompassing various categories. On the collected dataset,  we employ hashing techniques for deduplication ensuring the uniqueness of models. Additionally, we remove models with poor quality, which are excessively simplistic or intricate, as well as exceptionally rare instances. Following this, we classify these models based on ICS \cite{ICS} into 24 main categories. Ultimately, we obtain a clean dataset consisting of 926 models for experiments. 

\noindent{\bfseries Implementation Details } All experiments are conducted on the Ubuntu 20.04 operating system. Our hardware specifications include an Intel Xeon Gold 6326 CPU, 32GB RAM, and an NVIDIA GeForce RTX 4090. The batch size is set to 32. Contour sketches from Stage-one are processed to a size of 224 × 224 pixels. Detailed information about experiments is provided in the \textit{Appendix}.

\subsection{Qualitative Evaluation}\label{4.3}
Due to the absence of research on the same task, we intend to compare our approach  from two perspectives, which involve approaches designed for generating engineering sketches and existing state-of-the-art freehand sketches generative methods.

\begin{figure}[h] 
  \centering
  \includegraphics[width=\linewidth]{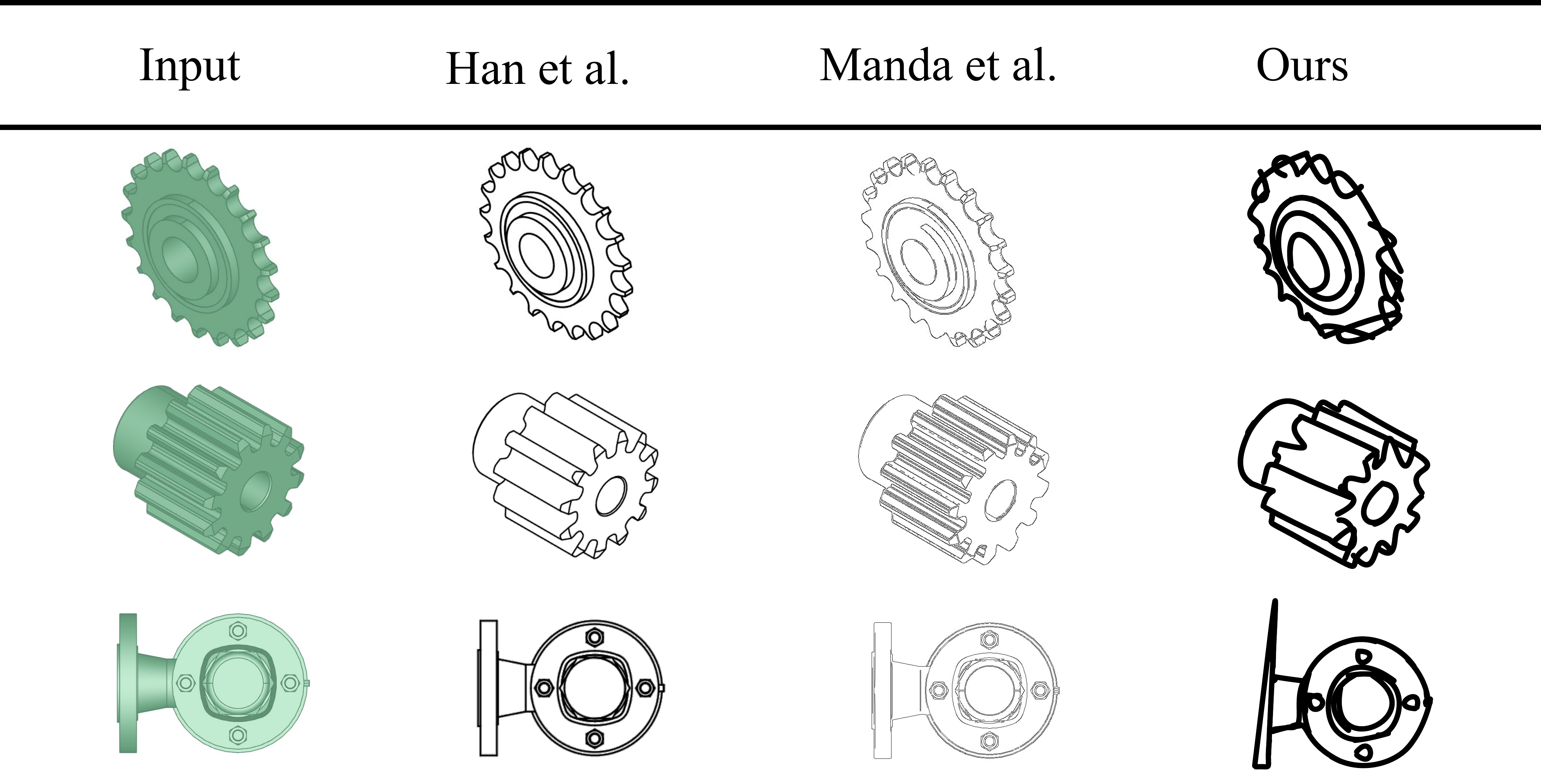}
  % \vspace{-0.5em}
  \caption{Comparison to other methods for generating sketches of mechanical components. }
  \label{fig:comparsion-tra}
\end{figure}

\noindent{\bfseries Sketches of mechanical components } In Figure \ref{fig:comparsion-tra} , we contrast our method with Han et al. \cite{han2020spare3d} and Manda et al. \cite{manda2021cadsketchnet} , using our collected components as inputs. Han et al. \cite{han2020spare3d} use PythonOCC \cite{paviot2018pythonocc} to produce view drawings, while Manda et al. \cite{manda2021cadsketchnet} create sketches through image-based edge extraction techniques. Although their results preserve plentiful engineering features, it is apparent that their outcomes resembling extracted outlines from models lack the style of freehand, which limits applicability in freehand sketch modeling. In contrast, our approach almost retains essential information of mechanical components equivalent to their results, such as through holes, gear tooth, slots, and overall recognizable features, while our results also demonstrate an excellent freehand style.

% \vspace{-0.5em}

\begin{figure}[h] 
  \centering
  \includegraphics[width=\linewidth]{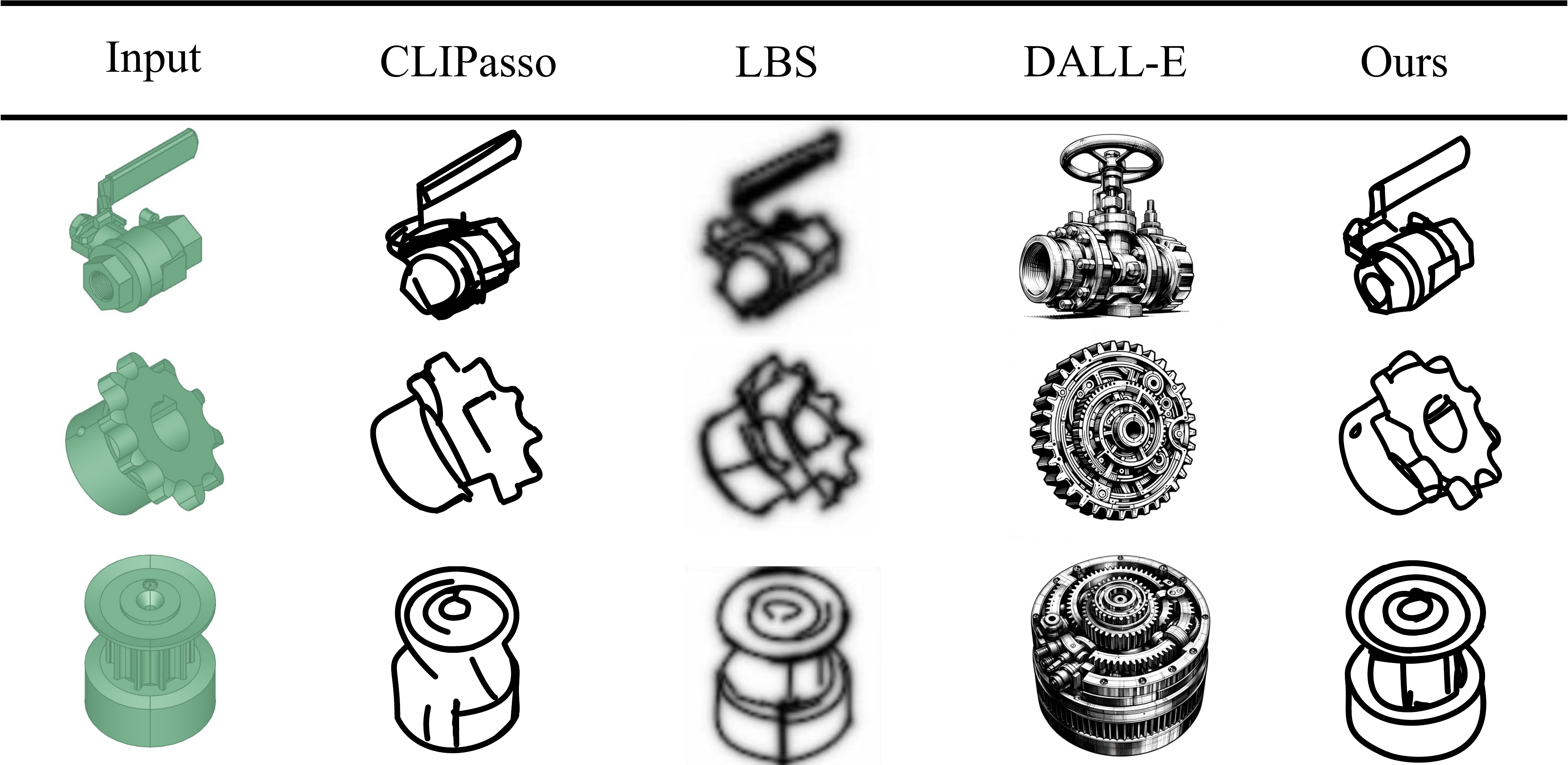}
  % \vspace{-1em}
  \caption{Comparison to other state-of-the-art method for generating sketches with a freehand style. }
  \label{fig:comparsion-abs}
\end{figure}

% \vspace{-0.75em}
\noindent{\bfseries Sketch with a freehand style } We compare our method with excellent freehand sketch generative methods like CLIPasso \cite{vinker2022clipasso} and LBS \cite{lee2023learning} . Moreover, we present a contrast with  DALL-E \cite{ramesh2021zero} , which is a mainstream large-model-based image generation approach. As shown in Figure \ref{fig:comparsion-abs} , all results are produced by 25 strokes using our collected dataset. In the first example, CLIPasso's \cite{vinker2022clipasso} result exhibits significant disorganized strokes, and LBS \cite{lee2023learning} almost completely covers the handle of valve with numerous strokes, leading to inaccurate representation of features. In the second and third examples, results by CLIPasso \cite{vinker2022clipasso} lose key features, such as the gear hole and the pulley grooves. For LBS \cite{lee2023learning} , unexpected stroke connections appear between modeling features and its stroke distribution is chaotic. In contrast, our strokes accurately and clearly are distributed over the features of components. These differences are attributed to the fact that CLIPasso \cite{vinker2022clipasso} initializes strokes via sampling randomly from the saliency map resulting in features that may not always be captured. Although LBS \cite{lee2023learning} modifies initialization of CLIPasso \cite{vinker2022clipasso} , it initializes strokes still relying on saliency maps influenced by noise information like monotonous colors and textures in mechanical components. Our method addresses this issue by introducing a novel edge-constraint initialization, which accurately places initial strokes on feature edges. Additionally, as LBS reported that its transformer-based model uses a CNN encoder. So its robustness comparison to our method will be similar to the results in Figure \ref{fig:comparsion-encoder} . In contrast to DALL-E \cite{ramesh2021zero} , we employ inputs consistent with previous experiments coupled with the prompt ("Create a pure white background abstract freehand sketch of input in 25 strokes") as the final inputs. It is evident that the large-model-based sketch generation method is still inadequate for our task.

% \vspace{-0.5em}

\begin{table*}[ht]
\centering
\caption{Quantitative comparison results by metrics. "-T" means test by transformed inputs. "-U" means test by unseen inputs. }
\label{tab:metrics evaluation}
\renewcommand\arraystretch{1.12}

\begin{tabular}{@{}l|cccccccccccc@{}}
\toprule
\multicolumn{1}{c|}{{}} & \multicolumn{4}{c}{Simple}                                                                                          & \multicolumn{4}{c}{Moderate}                                                                                       & \multicolumn{4}{c}{Complex}                                                                               \\ \cmidrule(l){2-13} 
\multicolumn{1}{c|}{Method}                       & \multicolumn{1}{l}{FID↓} & \multicolumn{1}{l}{GS↓} & \multicolumn{1}{l}{Prec↑} & \multicolumn{1}{l|}{Rec↑}          & \multicolumn{1}{l}{FID↓} & \multicolumn{1}{l}{GS↓} & \multicolumn{1}{l}{Prec↑} & \multicolumn{1}{l|}{Rec↑}          & \multicolumn{1}{l}{FID↓} & \multicolumn{1}{l}{GS↓} & \multicolumn{1}{l}{Prec↑} & \multicolumn{1}{l}{Rec↑} \\ \hline
CLIPasso \cite{vinker2022clipasso}                                    & 10.28                    & 5.70                    & 0.44                      & \multicolumn{1}{c|}{0.79}          & 12.03                    & 7.40                    & 0.35                      & \multicolumn{1}{c|}{0.72}          & 13.43                    & 9.91                    & 0.30                      & 0.69                     \\
LBS  \cite{lee2023learning}                                       & 9.46                     & 5.29                    & 0.45                      & \multicolumn{1}{c|}{0.81}          & 11.57                    & 7.03                    & 0.32                      & \multicolumn{1}{c|}{0.71}          & 12.71                    & 8.78                    & 0.31                      & 0.66                     \\
Ours                                        & \textbf{6.80}            & \textbf{3.37}           & \textbf{0.53}             & \multicolumn{1}{c|}{\textbf{0.87}} & \textbf{7.07}            & \textbf{3.96}           & \textbf{0.47}             & \multicolumn{1}{c|}{\textbf{0.83}} & \textbf{7.27}            & \textbf{4.52}           & \textbf{0.42}             & \textbf{0.81}            \\ \hline
Ours(VIT-B/32+adapter) -T                   & 7.01                     & 3.98                    & 0.48                      & \multicolumn{1}{c|}{0.83}          & 7.25                     & 6.08                    & 0.38                      & \multicolumn{1}{c|}{0.72}          & 7.42                     & 6.51                    & 0.32                      & 0.70                     \\
Ours(CNN) -T                                & 17.46                    & 28.10                   & 0.18                      & \multicolumn{1}{c|}{0.56}          & 19.44                    & 63.14                   & 0.13                      & \multicolumn{1}{c|}{0.37}          & 25.13                    & 79.44                   & 0.11                      & 0.25                     \\
Ours(VIT-B/32+adapter) -U                   & 8.60                     & 4.10                    & 0.44                      & \multicolumn{1}{c|}{0.81}          & 10.68                    & 6.20                    & 0.33                      & \multicolumn{1}{c|}{0.68}          & 13.44                    & 7.24                    & 0.31                      & 0.63                     \\
Ours(CNN) -U                                & 18.85                    & 30.33                   & 0.19                      & \multicolumn{1}{c|}{0.51}          & 20.78                    & 70.66                   & 0.11                      & \multicolumn{1}{c|}{0.40}          & 27.54                    & 87.54                   & 0.10                      & 0.20                     \\ \hline
\end{tabular}
\end{table*}

% \vspace{-1.5em}
\subsection{Quantitative Evaluation}\label{4.4}
{\bfseries Metrics Evaluation } We rasterize vector sketches into images and utilize evaluation metrics for image generation to assess the quality of  generated sketches. FID (Fréchet Inception Distance) \cite{heusel2017gans} quantifies the dissimilarity between generated sketches and standard data by evaluating the mean and variance of sketch features, which are extracted from Inception-V3 \cite{szegedy2016rethinking} pre-trained on ImageNet \cite{krizhevsky2012imagenet} . GS (Geometry Score) \cite{khrulkov2018geometry} is used to contrast the geometric information of data manifold between generated sketches and standard ones. Additionally,
we apply the improved precision and recall \cite{kynkaanniemi2019improved} as supplementary metrics following other generative works \cite{nichol2021improved} . In this experiment, we employ model outlines processed by PythonOCC \cite{paviot2018pythonocc} as standard data, which encapsulate the most comprehensive engineering information. The lower FID and GS scores and higher Prec and Rec scores indicate a greater degree of consistency in preserving modeling features between the generated sketches and the standard data. As shown in Table \ref{tab:metrics evaluation} , we classify generated sketches into three levels based on the number of strokes ($NoS$) : Simple ($16 \leq NoS \textless 24$ strokes), Moderate ($24 \leq NoS \textless 32$  strokes), and Complex ($32 \leq NoS \textless 40$  strokes). The first part of Table \ref{tab:metrics evaluation} showcases comparisons between our approach and other competitors, revealing superior FID, GS, Precision, and Recall scores across all three complexity levels. Consistent with the conclusions of qualitative evaluation, our approach retains more precise modeling features while generating freehand sketches. Additional metrics evaluation (standard data employ human-drawn sketches) is provided in the \textit{Appendix}.

\noindent{\bfseries User Study } We randomly select 592  mechanical components from 15 main categories in collected dataset as the test dataset utilized in user study. We compare results produced by Han et al. \cite{han2020spare3d} , Manda et al. \cite{manda2021cadsketchnet} , CLIPasso \cite{vinker2022clipasso} , LBS \cite{lee2023learning} and our method (the last three methods create sketches in 25 strokes). We invite 47 mechanical modeling researchers and ask them to score sketches based on two aspects:  engineering information and the freehand style. Scores range from 0 to 5, with higher scores indicating better performance in creating features and possessing a hand-drawn style. Finally, we compute average scores for all components in each method. As shown in Table \ref{tab:user study} , the result of  user study indicates that our method achieves the highest style score and overall score. These reveal our results have a human-prefer freehand style and  a better comprehensive performance in balancing information with style.

\begin{table}[h]
\centering
\caption{User study results. "Information" is the engineering information content score , "Style" denotes the score of freehand style, and "Overall" is the average of these two scores.}
% \vspace{-0.5em}
\label{tab:user study}
\renewcommand\arraystretch{1.15}
\begin{tabular}{l|ccc}
\hline
\multicolumn{1}{c|}{Method} & \multicolumn{1}{l}{Information↑} & \multicolumn{1}{l}{Style↑} & \multicolumn{1}{l}{Overall↑} \\ \hline
Han et al.\cite{han2020spare3d}                  & \textbf{4.20}                             & 0.84                       & 2.52                         \\
Manda et al.  \cite{manda2021cadsketchnet}              & 4.04                             & 1.21                       & 2.63                         \\
CLIPasso  \cite{vinker2022clipasso}                  & 2.71                             & 3.81                       & 3.26                         \\
LBS          \cite{lee2023learning}               & 2.94                             & 3.76                       & 3.35                          \\
\textbf{Ours}               & 3.80                             & \textbf{3.84}                       & \textbf{3.82}                \\ \hline
\end{tabular}
\end{table}

% \vspace{-1em}
\subsection{Performance of the Model } \label{4.2}
\begin{figure}[h] 
  \centering
  \includegraphics[width=\linewidth]{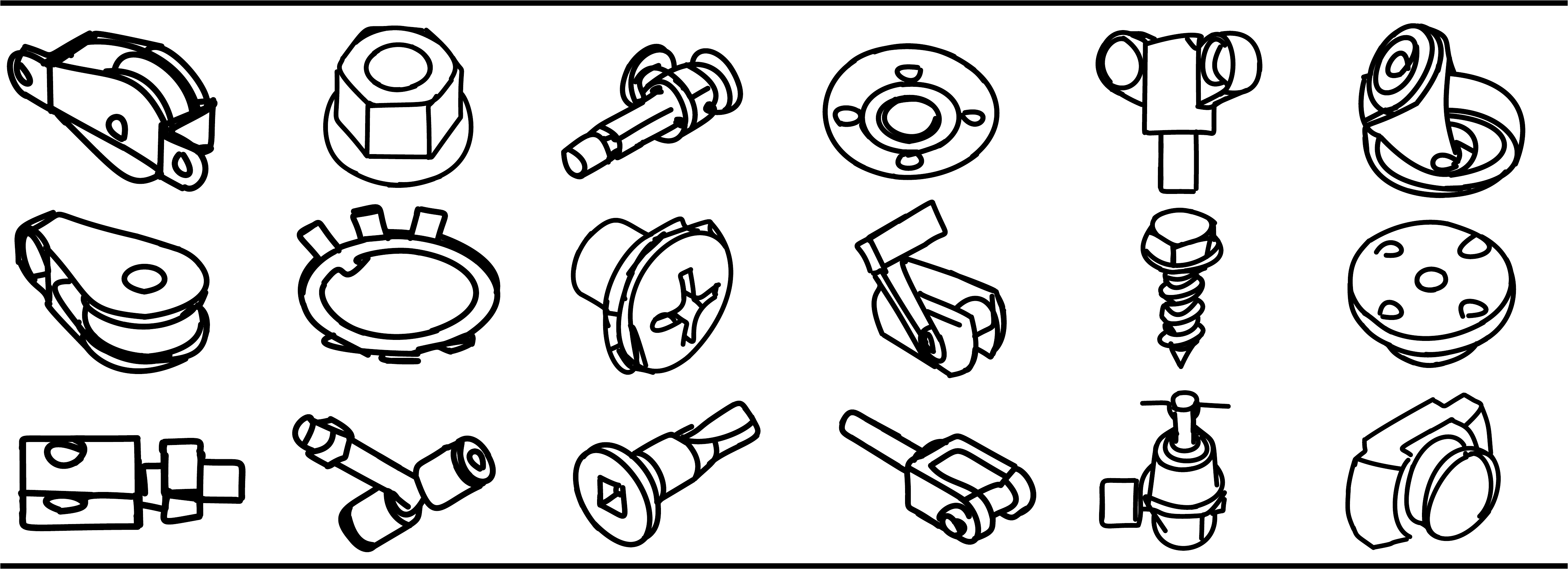}
  % \vspace{-2em}
  \caption{Robust performance across abundant categories. }
  \label{fig:variable}
\end{figure}

% {\bfseries Robustness } 
Different from traditional sketch generation methods, our generative model does not require additional sketch datasets. All training sketches are produced from our guidance sketch generator, which is optimized via CLIP \cite{CLIP} , a model pre-trained on four billion text-image pairs, producing high-quality guidance sketches. Benefiting from the guidance sketch generation process not being limited to specific categories, our method demonstrates robustness across a wide variety of mechanical components. In Figures \ref{fig:teaser} and \ref{fig:variable} , we showcase excellent generation results for various mechanical components. More qualitative results are provided in \textit{Appendix}.

% \vspace{-0.25em}

% \vspace{-1em}
\begin{figure}[h] 
  \centering
  \includegraphics[width=\linewidth]{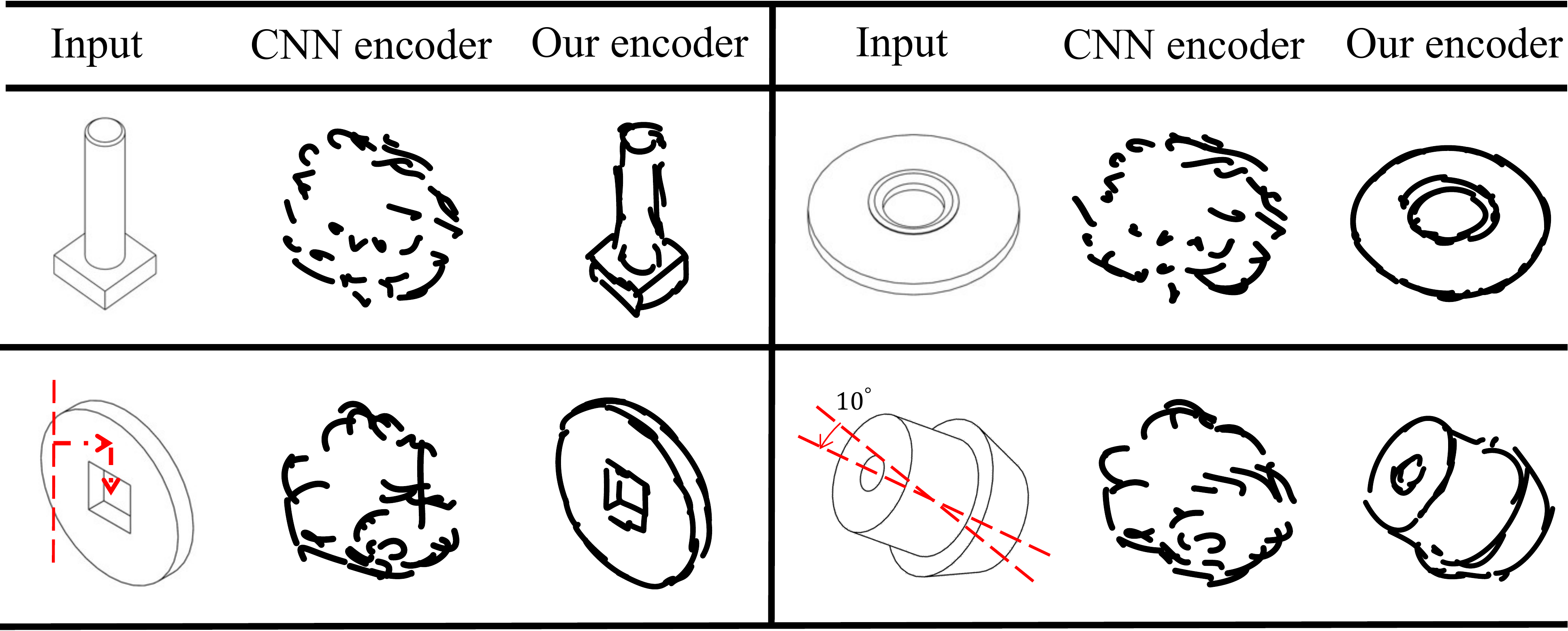}
  % \vspace{-2em}
  \caption{Comparison to our method with different encoders. }
  \label{fig:comparsion-encoder}
\end{figure}

\begin{table*}[ht]

\centering
\caption{Ablation Study with metrics evaluation. S-O : Stage-One, E-I : Guidance sketches generated by edge-constraint initialization , L-H: Training using  $ \mathcal{L}_{Hausdorff} $ . "-T" means test by transformed inputs. }
\label{tab:Ablation Study}. 
\renewcommand\arraystretch{1.2}
\begin{tabular}{l|ccc|cccccccccccc}
\hline
\multicolumn{1}{c|}{} & \multicolumn{1}{l}{}    & \multicolumn{1}{l}{}    & \multicolumn{1}{l|}{}    & \multicolumn{4}{c}{Simple}                                                                                          & \multicolumn{4}{c}{Moderate}                                                                                        & \multicolumn{4}{c}{Complex}                                                                               \\ \cline{5-16} 
Model                 & \multicolumn{1}{l}{S-O} & \multicolumn{1}{l}{I-O} & \multicolumn{1}{l|}{L-H} & \multicolumn{1}{l}{FID↓} & \multicolumn{1}{l}{GS↓} & \multicolumn{1}{l}{Prec↑} & \multicolumn{1}{l|}{Rec↑}          & \multicolumn{1}{l}{FID↓} & \multicolumn{1}{l}{GS↓} & \multicolumn{1}{l}{Prec↑} & \multicolumn{1}{l|}{Rec↑}          & \multicolumn{1}{l}{FID↓} & \multicolumn{1}{l}{GS↓} & \multicolumn{1}{l}{Prec↑} & \multicolumn{1}{l}{Rec↑} \\ \hline
Ours                  &                         & \ding{51}                       & \ding{51}                        & 9.01                     & 4.73                    & 0.45                      & \multicolumn{1}{c|}{0.81}          & 10.57                     & 6.79                    & 0.39                      & \multicolumn{1}{c|}{0.75}          & 11.11                    & 7.20                    & 0.31                      & 0.68                     \\
Ours                  & \ding{51}                       &                         & \ding{51}                        & 7.69                     & 4.38                    & 0.47                      & \multicolumn{1}{c|}{0.82}          & 8.28                     & 5.08                    & 0.40                      & \multicolumn{1}{c|}{0.78}          & 8.62                     & 6.43                    & 0.33                      & 0.70                     \\
\textbf{Ours}         & \ding{51}                       & \ding{51}                       & \ding{51}                        & \textbf{6.80}            & \textbf{3.37}           & \textbf{0.53}             & \multicolumn{1}{c|}{\textbf{0.87}} & \textbf{7.07}            & \textbf{3.96}           & \textbf{0.47}             & \multicolumn{1}{c|}{\textbf{0.83}} & \textbf{7.27}            & \textbf{4.52}           & \textbf{0.42}             & \textbf{0.81}            \\ \hline
Ours -T               & \ding{51}                       & \ding{51}                       &                          & 9.42                     & 5.38                    & 0.40                      & \multicolumn{1}{c|}{0.74}          & 10.23                     & 7.34                    & 0.32                      & \multicolumn{1}{c|}{0.65}          & 11.04                    & 8.77                    & 0.21                      & 0.63                     \\
Ours -T               & \ding{51}                       & \ding{51}                       & \ding{51}                        & 7.01                     & 3.98                    & 0.48                      & \multicolumn{1}{c|}{0.83}          & 7.25                     & 5.28                    & 0.38                      & \multicolumn{1}{c|}{0.72}          & 7.42                     & 6.51                    & 0.32                      & 0.70                     \\ \hline
\end{tabular}
\end{table*}

% \vspace{-0.5em}
Previous works like \cite{lee2023learning} predominantly employ a CNN encoder that uses fixed-size convolution kernels and pooling layers to extract local features. It leads to the neglect of global information, resulting in poor robustness. To address this issue, we utilize a CLIP ViT-B/32 combined with an adapter as our encoder. Qualitative and quantitative comparative experiments are designed to demonstrate the efficacy of our encoder. In the first row of Figure \ref{fig:comparsion-encoder} , we employ models which are similar-category, but unseen in training as test inputs. Compared to the method using a CNN encoder (ResNeXt18 \cite{xie2017aggregated} is used in this experiment), which only produces chaotic and shapeless strokes, the method with our encoder creates sketches with recognizable overall contours and essential features. In the second row, we apply contour sketches seen in training as inputs, each of which is transformed to the right and downward by 5 pixels and rotated counterclockwise by 10\degree. It can be observed that the method with our encoder still accurately infers component sketches, whereas the one using a CNN encoder fails to generate recognizable features. The quantitative comparison results are presented in the second part of Table \ref{tab:metrics evaluation} . Consistent with our expectations, the method with our encoder performs better in terms of evaluation metrics. It showcases that our encoder fortifies the encoding robustness for unseen and transformed inputs, enhancing the generalization and equivariance of the model.

\begin{figure}[h] 
  \centering
  \includegraphics[width=\linewidth]{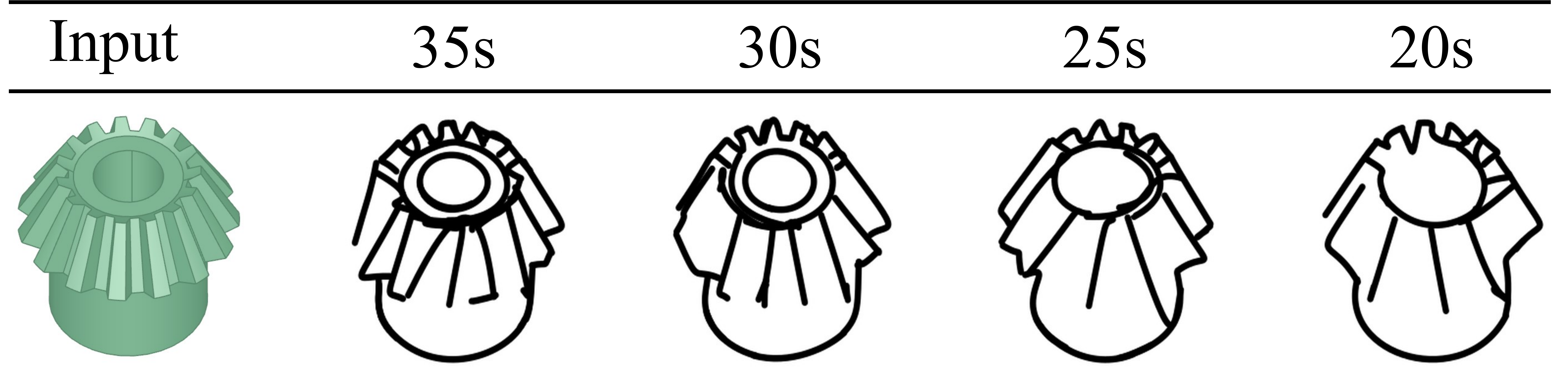}
  % \vspace{-2em}
  \caption{Different levels of abstraction generated by ours. Left to right: gear model  and results in 35, 30, 25, 20 strokes.}
  \label{fig:abstraction}
\end{figure}

Abstraction is an important characteristic of freehand sketches.
% simultaneously maintaining abstraction while preserving recognizability, underlying structure, and essential visual components of the subject has always been a challenge.
Our method effectively achieves it by individually training the stroke generator on different levels of abstraction sketches datasets. As shown in Figure \ref{fig:abstraction} , we respectively present the generated sketches from an input gear component using 35, 30, 25, and 20 strokes. As the number of strokes decreases, the abstraction level of gear sketches increases. Our method constrains strokes to create the essence of the gear. Iconic characteristics of a gear such as the general contour, teeth, and tooth spaces can be maintained, even though some minor details like through-holes may be removed.

% \vspace{-0.5em}

% \vspace{-1em}
\subsection{Ablation Study}\label{4.5}
\begin{figure}[h] 
  \centering
  \includegraphics[width=\linewidth]{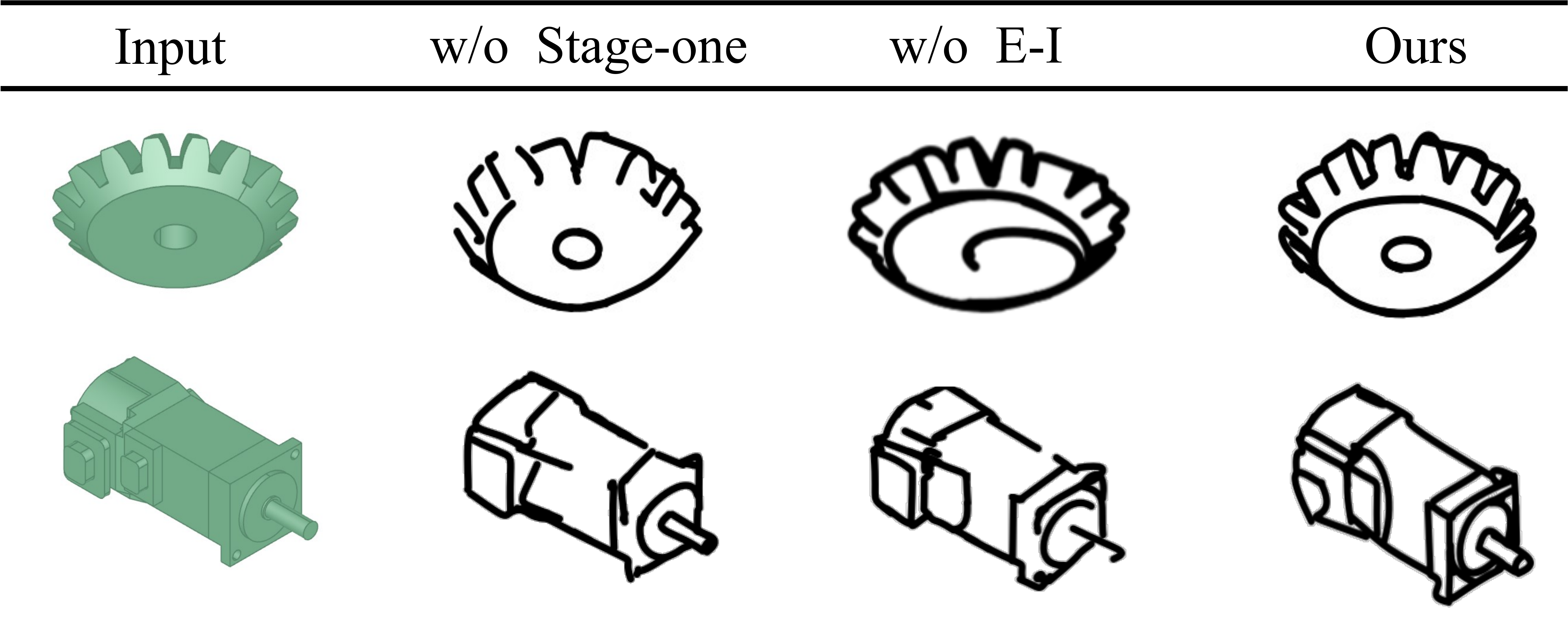}
  % \vspace{-2em}
  \caption{Ablation study. E-I: Edge-constraint initialization, "Ours" are the results produced by our complete framework. }
  \label{fig:ablation}
\end{figure}

{\bfseries Stage-One } As shown in Figure \ref{fig:ablation} , the results of the method lacking Stage-One are susceptible to issues such as  producing unstructured features and line distortions in qualitative ablation experiment. Excellent metric scores in Table \ref{tab:Ablation Study} demonstrate our complete framework can create richer and more accurate modeling information. This improvement is attributed to Stage-One, which filters out noise information such as color, texture, and shadows, mitigating their interference with the generation process.

\noindent{\bfseries Edge-constraint Initialization } In order to verify whether edge-constraint initialization can make precise geometric modeling features, we remove the optimized mechanism in the initial process. Comparison in Figure \ref{fig:ablation} clearly demonstrates that sketches generated with edge-constraint initialization(E-I) exhibit better performance in details generation and more reasonable stroke distribution. These benefit from E-I ensuring that initial strokes are accurately distributed on the edges of model features. Similarly, we utilized quantitative metrics to measure the generation performance. As shown in Table \ref{tab:Ablation Study} , sketches generated after initialization optimization achieve improvements in metrics such as FID, GS, and so on.

\noindent{\bfseries Hausdorff distance Loss } \textit{Hausdorff distance} is a metric used to measure the distance between two shapes, considering not only the spatial positions but also the structural relationships between shapes. By learning shape invariance and semantic features, the model can more accurately match shapes with different transformations and morphologies, aiding in the model's equivariance. The ablation experimental result is depicted in Table \ref{tab:Ablation Study} . It is evident that all the quantitative metrics for our method training with Hausdorff distance become better on the transformed test dataset. 

% \vspace{-0.5em}

% \vspace{-1em}
\section{Conclusion and Future Work}
This paper proposes a novel two-stage framework, which is the first time to generate freehand sketches for mechanical components. We mimic the human sketching behavior pattern that produces optimal-view contour sketches in Stage-One and then translate them into freehand sketches in Stage-Two. To retain abundant and precise modeling features, we introduce an innovative edge-constraint initialization. Additionally, we utilize a CLIP vision encoder and propose a \textit{Hausdorff distance}-based guidance loss to improve the robustness of the model. Our approach aims to promote research on data-driven algorithms in the freehand sketch domain. Extensive experiments demonstrate that our approach performs superiorly compared to state-of-the-art methods. 

Through experiments, we discover that we would better utilize a comprehensive model rather than direct inference to obtain desirable outcomes for unseen models with significant geometric differences. In future work, we will explore methods to address this issue, further enhancing the model's generalizability.

\begin{acks}

This work is partially supported by National Key Research and Development Program of China (2022YFB3303101). We thank \href{https://orcid.org/0009-0003-5190-7104}{Chaoran Zhang}, \href{https://orcid.org/0009-0007-7480-7770}{Fuhao Li}, Hao Zhang, and Duc Vu Minh for their valuable advice, which helped to improve this work. (\href{https://orcid.org/0009-0003-5190-7104}{Chaoran Zhang}, \href{https://orcid.org/0009-0007-7480-7770}{Fuhao Li}, and Hao Zhang are members of the Tsinghua IMMV Lab.) We also thank \href{https://www.newdimchina.com/}{Newdim} for providing data support.

\end{acks}

\bibliographystyle{ACM-Reference-Format}
\bibliography{sample-sigconf}

\clearpage

\appendix









\section{Initialization Analysis}
In this section, we will meticulously contrast and analyze our Edge-constant initialization with the original CLIPasso \cite{vinker2022clipasso} initialization method.

As described in CLIPasso \cite{vinker2022clipasso}, it utilizes the ViT-32/B CLIP \cite{CLIP} to obtain the salient regions of a target image. This is achieved by averaging the attention outputs from all attention heads across each self-attention layer, generating a total of 12 attention maps. These maps are further averaged to derive the relevancy map, obtained by examining the attention between the final class embedding and all 49 patches. Subsequently, this relevancy map is combined with the edge map obtained through XDoG \cite{winnemoller2012xdog} extraction. The resulting attention map is then utilized to determine the locations for the initial strokes. In the process of determining the initial positions of strokes, CLIPasso\cite{vinker2022clipasso} utilizes random seeds on the saliency map to sample positions for the first control point of each curve. Following this, it randomly selects the subsequent three control points of each Bezier curve within a small radius (0.05) of the initial point.

Such random initialization methods often result in the initial points of strokes being inadequately distributed around the critical features of mechanical components during sketch generation, leading to the loss of substantial modeling information. Moreover, this approach frequently results in an excessive placement of initial stroke points in certain prominent feature areas, causing confusion in generating sketch strokes and preventing accurate representation of modeling features. To address this issue, we propose the edge-constant initialization to deterministically sample. We utilize SAM \cite{kirillov2023segment} to perform feature segmentation on the input contour sketch. Based on the segmentation results, we predefine four stroke initialization points evenly spaced along the edge of each segmented feature. Subsequently, we dynamically change the initialization points based on the comparison with the manually required number of generated strokes. If the requested number of strokes is less than the total predefined initialization points, we evenly discard points contained within each segmented feature. Conversely, if the requested number of strokes exceeds the total predefined initialization points, we employ a greedy algorithm on the saliency map of the target image to determine additional stroke initialization points in the most salient regions \cite{lee2023learning}. This initialization method not only ensures the precise generation of mechanical component features but also optimizes the distribution of generated strokes, resulting in clearer generated sketches.

\begin{figure}[h]
  \centering
  \includegraphics[width=\linewidth]{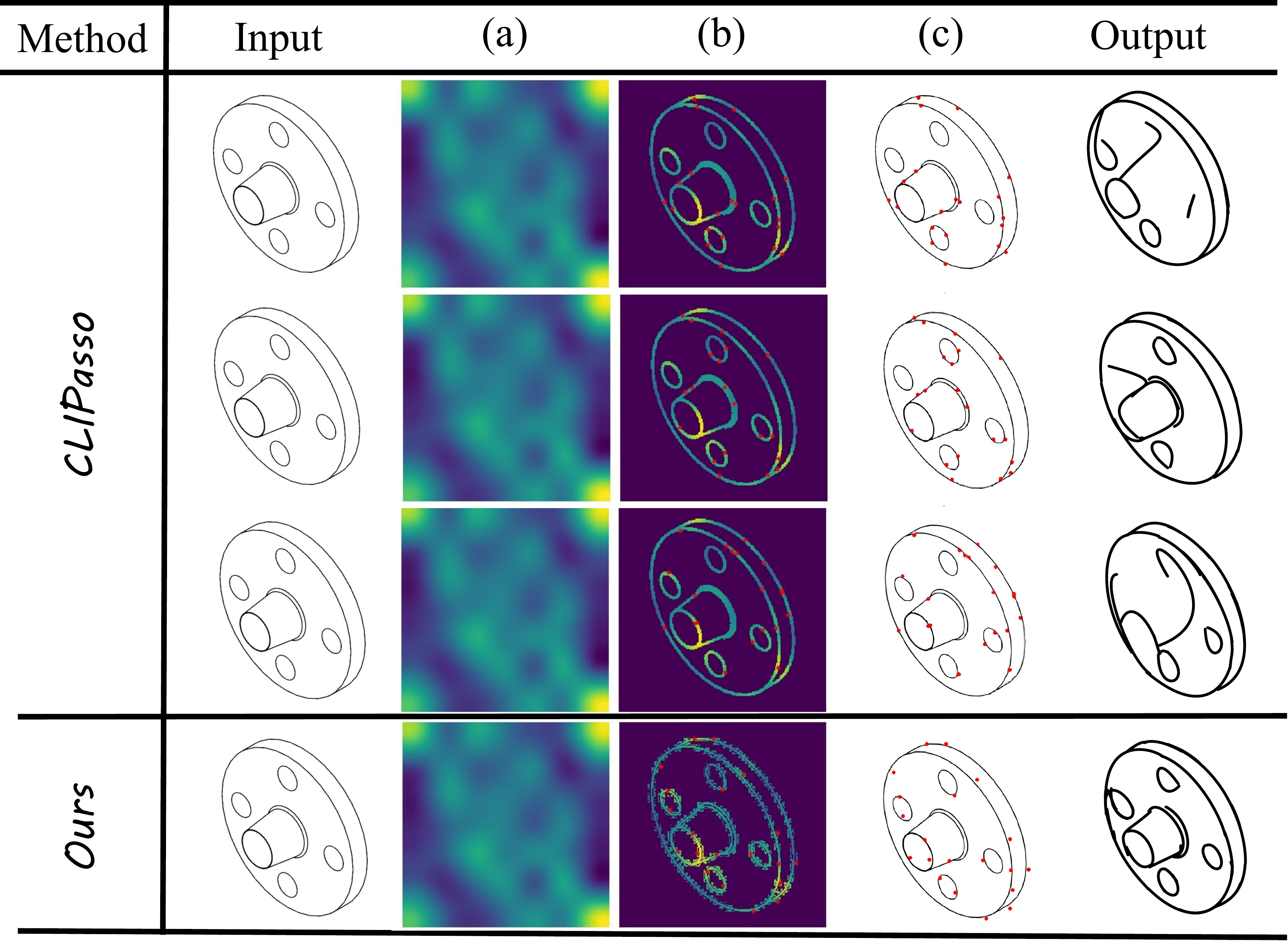}
  \caption{Strokes Initialization. All sketches are produced with 20 strokes. Left to right: input contour sketches, (a) the saliency maps generated from CLIP ViT activations, (b) and (c)  are initial stroke locations (in red) in final distribution maps and inputs,  output freehand sketches.}
  \Description{}
  \label{fig:Initialization Analysis}
\end{figure}

As shown in Figure \ref{fig:Initialization Analysis}, we conduct experiments using three sampling strategies of random seeds provided by CLIPasso\cite{vinker2022clipasso}. It is evident that in the first instance, no stroke initialization points are placed at the three through-holes of the input flange contour sketch, resulting in the loss of this important feature in the result. In the second instance, the placement of three stroke initialization points on the upper right through-hole is unnecessary, as it is a simple feature that does not require three strokes to depict. In the third instance, three stroke initialization points are clustered around the edge contour of the flange, while only two initialization points are placed on the structurally complex cylindrical section. These illustrate the irrational distribution of stroke initialization points caused by the random seed sampling method, ultimately leading to unsatisfactory sketch generation results. In contrast, our proposed edge-constant initialization optimizes the placement of stroke initialization points, ensuring their rational distribution on modeling feature edges. It can be observed that sketches generated through our improved method adequately preserve crucial modeling features, with a clear and rational distribution of strokes.

\section{Stability Analysis}
In this section,  we  will evaluate the stability of our transformer-based \cite{Ribeiro_2020_CVPR,Liu_2021_ICCV,lee2023learning}  stroke generator.

In Stage-Two, our improved initialization method has enhanced the guidance sketch generator to produce informative freehand sketches. However, the guidance sketch generator employs an optimizer to create sketches through thousands of optimization iterations during sketch generation, leading to uncertainty in the outcomes. Each step of this optimization-based process is guided by CLIP \cite{CLIP} in terms of both semantic and geometric similarities to create strokes. This optimization process is uncontrollable and the optimized result from each step exhibits variability. It results in unstable and uncontrollable quality performance of the generated sketches. In order to consistently generate high-quality sketches, we adopt a transformer-based \cite{Ribeiro_2020_CVPR,Liu_2021_ICCV,lee2023learning} generative framework. We extract intermediate sketches from the optimization process of the guidance sketch generator as ideal guides for process sketches from each intermediate layer in the stroke generator. we utilize guidance loss during training to ensure that the stroke generator learns  features from corresponding  intermediate process guidance sketches. We employ CLIP-based \cite{CLIP} perceptual loss to ensure the similarity between the generated freehand sketches and contour sketches in both geometry and semantic information. Through training, all learned features are fixed into determined weights. During the inference phase, our model can rapidly infer freehand sketches based on the trained weights. This generation approach ensures output consistency and achieves satisfactory generation quality.

\begin{figure*}[hb]
  \centering
  \includegraphics[width=0.8\textwidth]{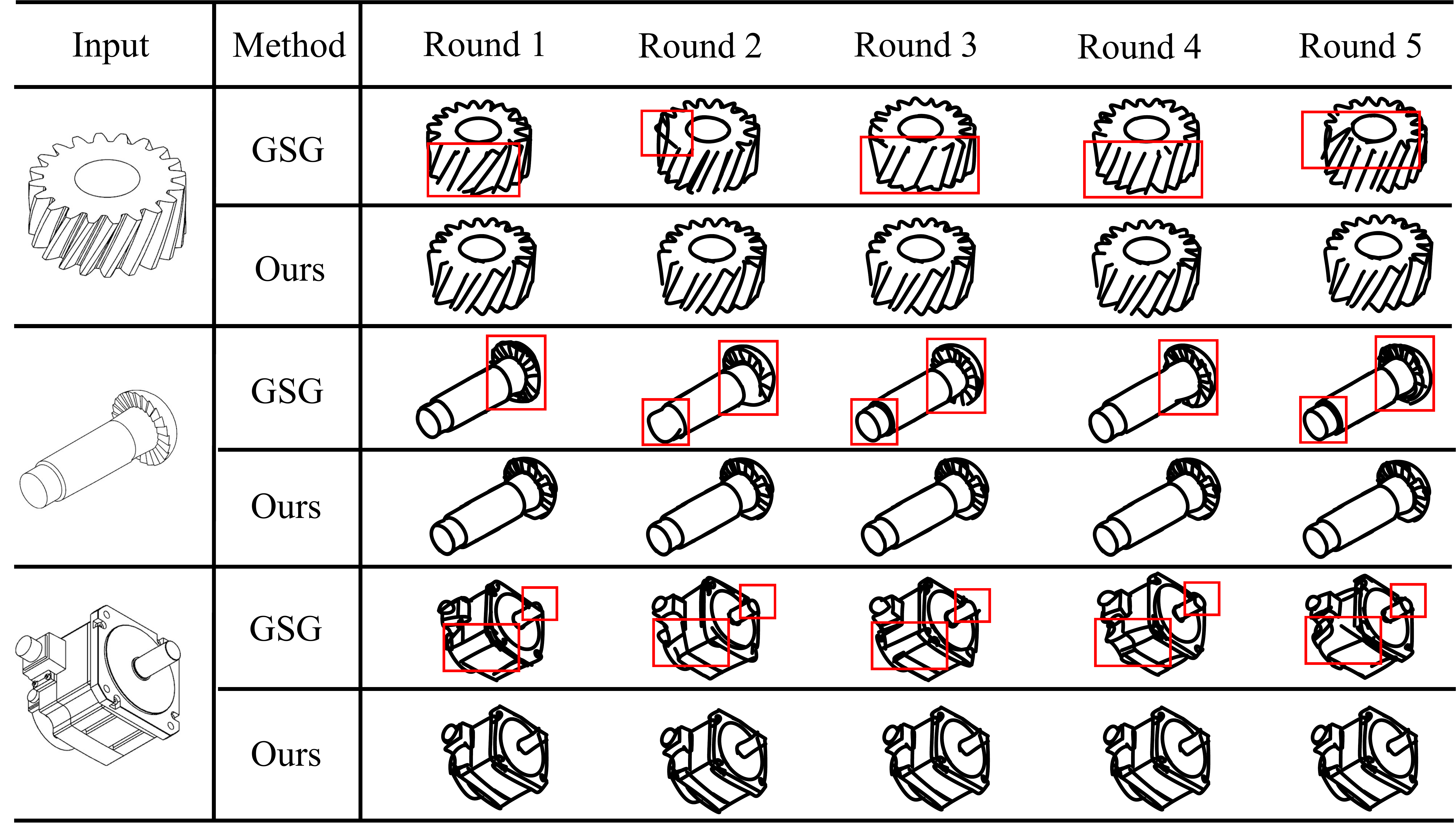}
  \caption{Stability analysis for our generative framework. All sketches are produced with 20 strokes. "GSG" refers to the method of directly generating sketches through the guidance sketch generator, while "ours" represents the method of inferring sketches by our trained complete framework. The region marked by the red rectangle represents the area of significant variation in the sketches generated by the GSG. }
  \Description{}
  \label{fig:Stability Analysis}
\end{figure*}

We design comparative experiments to validate the numerical stability of our generative framework. Using the same inputs, we conduct five rounds of sketch generation experiments separately with only the guidance sketch generator (GSG) and the complete generative framework (trained on the collected mechanical component dataset). As shown in  Figure \ref{fig:Stability Analysis}, the outcomes produced by the guidance sketch generator (GSG) for mechanical freehand sketches vary each time, and some of them exhibit suboptimal performance. For instance, in the case of the first instance, the distribution of the gear teeth slots varies significantly in each generated result, and due to the instability of the optimization-based generation method, issues arise such as chaotic stroke composition in the second round's results and erroneous connections between teeth slot strokes and through-hole strokes in the fifth round's generated sketches. Similar situations are also evident in the second and third instances. For example, in the second instance, the distribution of continuous sections of the flat-head screws differs in each round of the experiment. And results occasionally are accompanied by  contour loss such as the loss of the bottom circle of the screw in the second round of experiments, and the loss of connection at the head of the screw in the fourth round. In the third instance, involving a complex motor model, the strokes creating the main body of the motor within the area marked by the red rectangle exhibit significant variations in distribution across each experimental round. Additionally, some results accurately depict small through-holes on the motor surface, while others fail to capture this information. The reason for these issues arises from the uncontrollable nature of the optimization-based generation process. Despite our efforts to accurately position stroke initialization points on features during preprocessing, deviations in geometric and semantic guidance during optimization may result in inadequate representations of certain details in the generated sketches. In contrast, our comprehensive generation framework, after being trained on a large and diverse dataset of mechanical components, fixes learned features into weights. This ensures consistent outputs in each round of testing, and stable representations of modeling features for the components.

\clearpage
\section{Implementation Details}
In order to tailor our method specifically for freehand  sketch generation in engineering freehand sketch modeling, we build a CAD dataset exclusively comprising mechanical components in the STEP format. we invite numerous mechanical modeling  researchers to collect mechanical components from the TraceParts \cite{TraceParts}. They are asked to encompass a diverse array of categories to enhance the inference generalization of our generative model. In the end, we obtain a Raw dataset including nearly 2,000 mechanical components.
For the collected raw dataset, we employ hashing techniques for deduplication, ensuring the uniqueness of models in the dataset. Subsequently, we remove models with poor quality, which are excessively simplistic or intricate, as well as exceptionally rare instances. Following this, we classify these models based on the International Classification for Standards (ICS) \cite{ICS} into 24 main categories, comprising 180 corresponding subcategories. Ultimately, we obtained a clean dataset consisting of 926 models.

We implement the methods of Stage One using Python3 with PythonOCC and PyTorch, where PyTorch supports the viewpoint selector. For Stage Two, PyTorch and DiffVG are used to implement the model, where DiffVG is used for the differentiable rasterizer.

% \usepackage{multirow}
\begin{table*}[ht]
\centering
\caption{Quantitative comparison results by metrics using real human-drawn sketches as standard data.}
\label{tab:additional metrics evaluation}
\renewcommand\arraystretch{1.5}
\begin{tabular}{l|cccccccccccc}
\hline
\multicolumn{1}{c|}{{}}& \multicolumn{4}{c}{Simple}                                                                                          & \multicolumn{4}{c}{Moderate}                                                                                        & \multicolumn{4}{c}{Complex}                                                                               \\ \cline{2-13} 
\multicolumn{1}{c|}{Method}                 & \multicolumn{1}{l}{FID↓} & \multicolumn{1}{l}{GS↓} & \multicolumn{1}{l}{Prec↑} & \multicolumn{1}{l|}{Rec↑}          & \multicolumn{1}{l}{FID↓} & \multicolumn{1}{l}{GS↓} & \multicolumn{1}{l}{Prec↑} & \multicolumn{1}{l|}{Rec↑}          & \multicolumn{1}{l}{FID↓} & \multicolumn{1}{l}{GS↓} & \multicolumn{1}{l}{Prec↑} & \multicolumn{1}{l}{Rec↑} \\ \hline
Han et al. \cite{han2020spare3d}                                 & 12.66                    & 9.54                    & 0.48                      & \multicolumn{1}{c|}{0.75}          & 13.83                    & 11.44                   & 0.39                      & \multicolumn{1}{c|}{0.69}          & 14.68                    & 17.22                   & 0.35                      & 0.68                     \\
Manda et al.    \cite{manda2021cadsketchnet}                            & 14.51                    & 9.71                    & 0.47                      & \multicolumn{1}{c|}{0.74}          & 15.17                    & 12.73                   & 0.41                      & \multicolumn{1}{c|}{0.70}          & 15.43                    & 18.80                   & 0.33                      & 0.66                     \\
CLIPasso \cite{vinker2022clipasso}                                   & 12.75                    & 7.59                    & 0.42                      & \multicolumn{1}{c|}{0.71}          & 13.67                    & 10.61                   & 0.32                      & \multicolumn{1}{c|}{0.67}          & 14.51                    & 13.94                   & 0.29                      & 0.65                     \\
LBS  \cite{lee2023learning}                                       & 12.40                    & 7.53                    & 0.43                      & \multicolumn{1}{c|}{0.73}          & 13.19                    & 9.24                    & 0.30                      & \multicolumn{1}{c|}{0.65}          & 14.03                    & 12.60                   & 0.28                      & 0.63                     \\
Ours                                        & \textbf{8.35}            & \textbf{4.77}           & \textbf{0.51}             & \multicolumn{1}{c|}{\textbf{0.83}} & \textbf{8.83}            & \textbf{5.43}           & \textbf{0.46}             & \multicolumn{1}{c|}{\textbf{0.81}} & \textbf{9.26}           & \textbf{6.57}           & \textbf{0.40}             & \textbf{0.78}            \\ \hline
\end{tabular}
\end{table*}

\section{Additional Quantitative Evaluation}

{\bfseries Metrics Evaluation } In section 4.3 of this paper, we employ evaluation metrics for image generation to assess the quality of generated sketches. Given the absence of benchmark datasets specifically for mechanical component  freehand sketches within the sketch community, we utilize component outlines processed through PythonOCC \cite{paviot2018pythonocc}, which encapsulate the most comprehensive engineering information, as the standard data. The experiment results demonstrate the superiority of our method over existing freehand sketch generation methods in preserving the modeling features of mechanical components. In this section, we will conduct quantitative metric evaluations on our method and other competitors using real human-drawn sketches of mechanical components collected by ourselves.

\begin{figure}[h]
  \centering
  \includegraphics[width=1.05\linewidth]{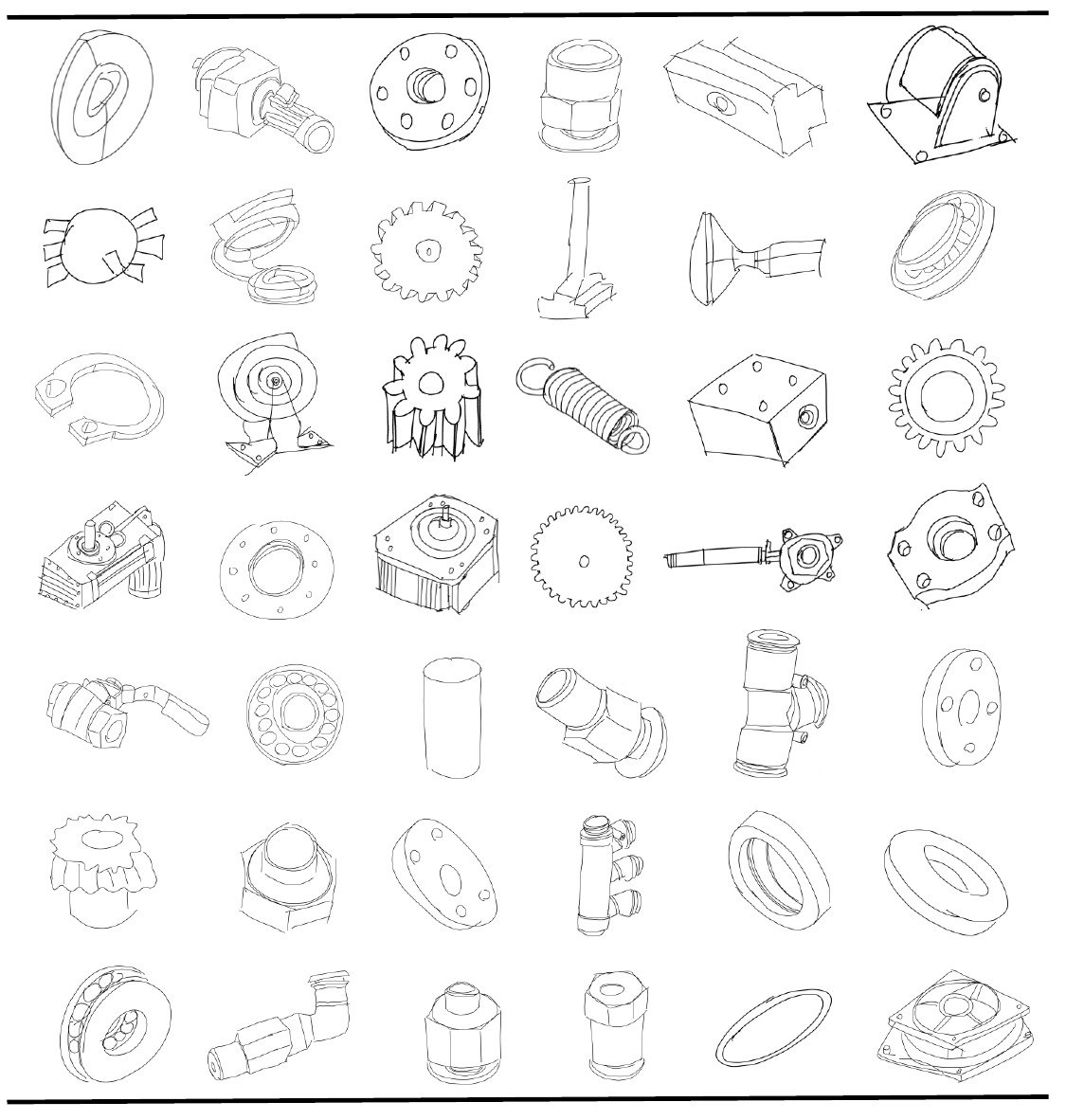}
  \caption{Our collection of real human-drawn sketches for mechanical components.
  }
  \Description{}
  \label{fig:Additional Quantitative Evaluation}
\end{figure}

We firstly introduce the construction process of the real human-drawn sketch dataset of mechanical component. From our collection of 926 three-dimensional mechanical component dataset, we randomly select 500 components. We invite 58 researchers with sketching expertise in the mechanical modeling domain, requesting them to draw a sketch for each component from a given perspective. We then obtain a test dataset comprising 500 mechanical sketches drawn by human engineers. As shown in the Figure \ref{fig:Additional Quantitative Evaluation}, we showcase the collection of real human-drawn sketches. Sketches of components drawn by researchers in the mechanical modeling domain preserve crucial modeling features which are essential for freehand sketch modeling. Correspondingly, certain minor details for modeling may be simplified, or overlooked by the researchers and not drawn. Moreover, it is evident that sketches crafted by humans exhibit a distinctive freehand style.

In this experiment, we utilize real human-drawn sketches of mechanical components as benchmark data, which balances maintaining crucial modeling features with exceptional freehand style well. We employ the same 500 components which are randomly selected during the construction of our real human-drawn sketches dataset as a test dataset for this experiment. Consistent with our previous approach, we generate sketches for components using different strokes based on their complexity, and categorize the generated sketches into three levels according to the number of strokes ($NoS$): Simple ($16 \leq NoS < 24$ strokes), Moderate ($24 \leq NoS < 32$ strokes), and Complex ($32 \leq NoS < 40$ strokes). We continue to evaluate the generated sketches using metrics such as FID, GS, and so on. As shown in Table \ref{tab:additional metrics evaluation}, we compare our method with  methods designed for generating engineering sketches as well as methods for producing freehand sketches. It is evident that our generation method achieved the most favorable metric scores across three different levels of complexity, demonstrating  the superiority of our approach in generating freehand sketches for mechanical components. In the experimental results, our outcomes obtain lower FID and GS scores and higher Prec and Rec scores. It indicates that our sketches more closely resemble real human-drawn sketches, exhibiting a higher level of consistency in preserving key modeling features and maintaining the freehand style between our results and real ones.

{\bfseries  Details for User Study. }  In the user study conducted in this paper, we invited 47 mechanical modeling researchers to rate the generated mechanical component sketches based on two dimensions: "information" and "style." In this part, we provide detailed explanations of the specific criteria represented by these two dimensions. In the "information" dimension, we ask the researchers to evaluate the completeness of modeling features contained in the sketches. This means that the higher the number of accurate modeling features retained in the generated sketches, the higher the score obtained. In the "style" dimension, we ask the researchers to assess the overall hand-drawn style of the sketches. Specifically, they were required to consider whether the generated sketches exhibit a hand-drawn style, whether the distribution of strokes in the generated sketches is reasonable, and whether it is more similar to the distribution structure of strokes drawn by humans.

From the results of the user study, it can be observed that Han et al. \cite{han2020spare3d} and Manda et al. \cite{manda2021cadsketchnet} perform better in the "information" dimension. This is because their sketches are generated by contour extraction from components, nearly retaining all modeling features. However, it is worth emphasizing that, to meet the requirements of improving modeling efficiency and lowering the modeling threshold, sketches used for freehand sketch modeling should mimic human-drawn characteristics as closely as possible that preserving key modeling features while simplifying or disregarding minor ones. Therefore, although Han et al. \cite{han2020spare3d} and Manda et al. \cite{manda2021cadsketchnet} retain relatively comprehensive features, they fail to meet the data requirements for freehand sketch modeling and they results lack a hand-drawn style, which fundamentally does not align with the demands of the task. Meanwhile, It can be observed that our generation results outperform  in preserving key features of  modeling among methods for generating freehand sketches. It demonstrates the effectiveness of the modules designed in our framework to retain crucial features. In terms of "style" dimension, our sketches perform best because they exhibit a hand-drawn style while maintaining a more reasonable stroke distribution, resembling the stroke distribution habits of human drawings. Considering both dimensions, our method achieved the highest overall scores, indicating that our approach performs better than existing methods in balancing the retention of key component modeling features and mimicking human hand-drawn style.

\section{Additional Qualitative Results}
\begin{figure*}[hb]
  \centering
  \includegraphics[width=0.95\textwidth,height=1.1\textwidth]{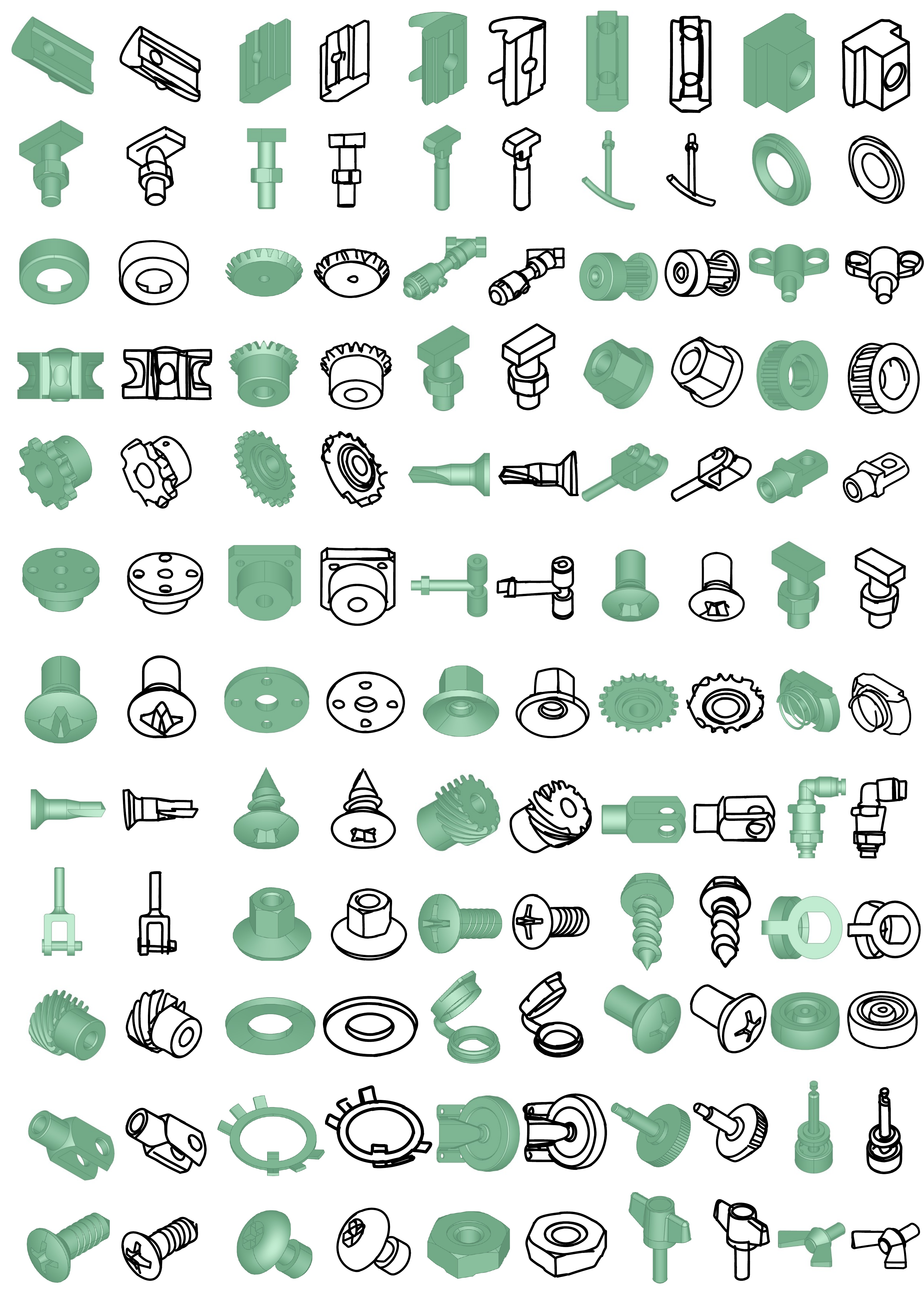}
  \caption{Robust performance across abundant categories.}
  \Description{}
  \label{fig:add-quali1}
\end{figure*}

\begin{figure*}[hb]
  \centering
  \includegraphics[width=\textwidth]{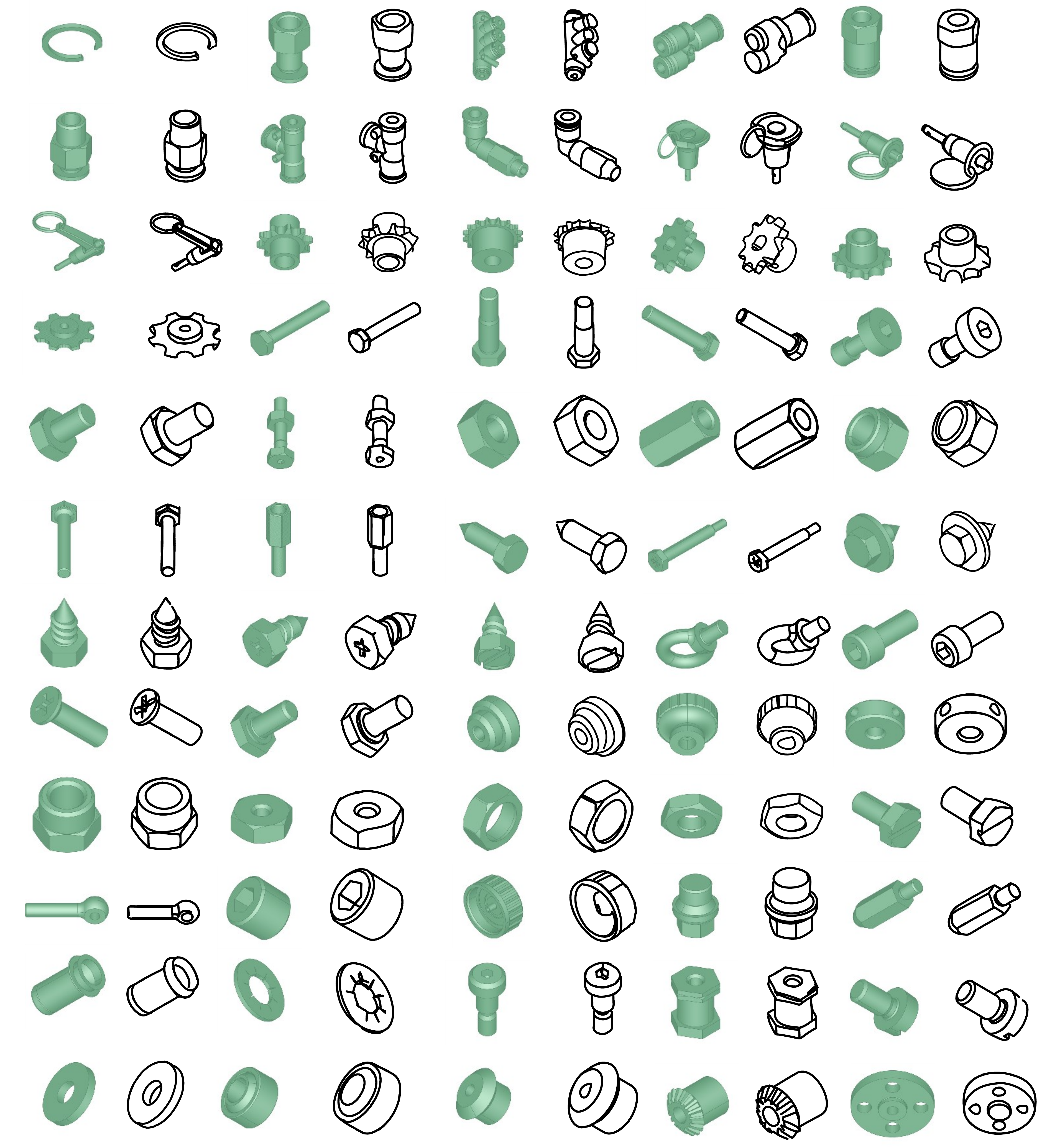}
  \caption{Robust performance across abundant categories.}
  \Description{}
  \label{fig:add-quali2}
\end{figure*}

\begin{figure*}[hb]
  \centering
  \includegraphics[width=\textwidth]{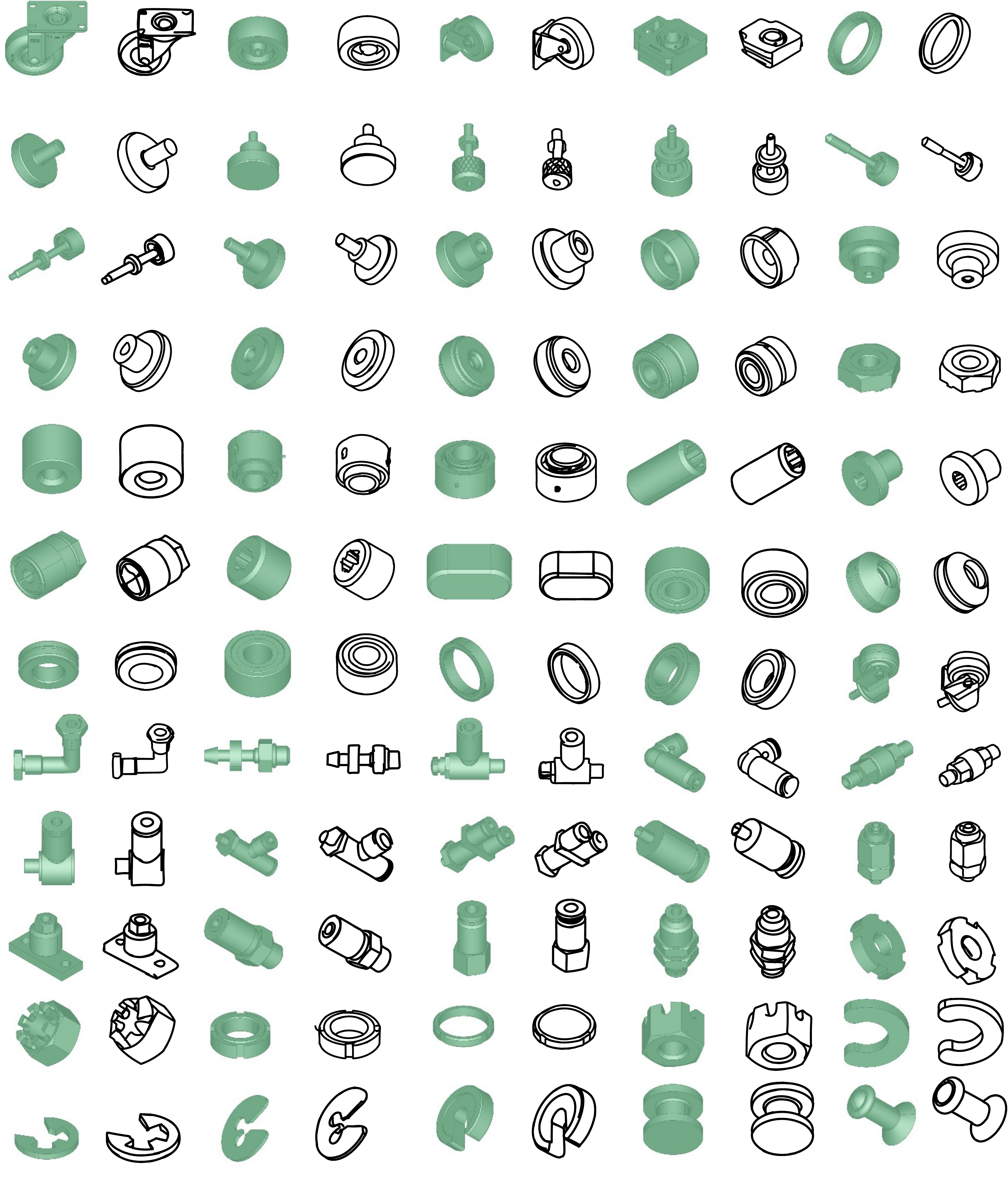}
  \caption{Robust performance across abundant categories.}
  \Description{}
  \label{fig:add-quali3}
\end{figure*}

Figure \ref{fig:add-quali1}, Figure \ref{fig:add-quali2}, and Figure \ref{fig:add-quali3} show a large number of excellent freehand sketches of mechanical components generated by our method.

\clearpage

\clearpage

\clearpage

\clearpage

\end{document}